\documentclass[10pt,twocolumn,letterpaper]{article}

\usepackage{cvpr}
\usepackage{times}
\usepackage{epsfig}
\usepackage{graphicx}
\usepackage{amsmath}
\usepackage{amssymb}
\usepackage{color}
\usepackage{multirow}
\usepackage{array}
\usepackage{caption}
\usepackage{subcaption}

\definecolor{dgreen}{rgb}{0,.7,0}
\definecolor{dyellow}{rgb}{.7,.7,0}
\definecolor{dred}{rgb}{.7,0,0}
\definecolor{dblue}{rgb}{0,0,0.7}

\definecolor{alexey}{rgb}{0.7,0,1}

\newcommand{\conv}{\textsc{conv}}

\newcommand{\fc}{\textsc{fc}}

\newcommand{\ds}{d}
\newcommand{\siftmap}{\mathbf{F}}
\newcommand{\siftfeat}{\mathbf{f}}
\newcommand{\ceil}[1]{\lceil #1 \rceil}
\newcommand{\st}{\!\times\!}
\newcommand{\ex}{\mathbb{E}}
\newcommand{\fnet}{f}

\newcommand{\feat}{\boldsymbol\phi}
\newcommand{\img}{\mathbf{x}}
\newcommand{\weights}{\mathbf{w}}

\graphicspath{{./images/}}

\usepackage[pagebackref=true,breaklinks=true,letterpaper=true,colorlinks,bookmarks=false]{hyperref}

\cvprfinalcopy 

\def\cvprPaperID{1892} 

\ifcvprfinal\fi
\begin{document}

\title{Inverting Visual Representations with Convolutional Networks}

\author{Alexey Dosovitskiy \qquad \qquad Thomas Brox\vspace{0.1cm}\\ 
University of Freiburg\\
Freiburg im Breisgau, Germany\\
{\tt\small \{dosovits,brox\}@cs.uni-freiburg.de}
}

\maketitle

\begin{abstract}
Feature representations, both hand-designed and learned ones, are often hard to analyze and interpret, even when they are extracted from visual data.
We propose a new approach to study image representations by inverting them with an up-convolutional neural network.
We apply the method to shallow representations (HOG, SIFT, LBP), as well as to deep networks.
For shallow representations our approach provides significantly better reconstructions than existing methods, revealing that there is surprisingly rich information contained in these features.
Inverting a deep network trained on ImageNet provides several insights into the properties of the feature representation learned by the network.
Most strikingly, the colors and the rough contours of an image can be reconstructed from activations in higher network layers and even from the predicted class probabilities.
\end{abstract}

\section{Introduction}
A feature representation useful for pattern recognition tasks is expected to concentrate on properties of the input image which are important for the task and ignore the irrelevant properties of the input image.
For example, hand-designed descriptors such as HOG~\cite{DalalTriggs_2005} or SIFT~\cite{Lowe_2004}, explicitly discard the absolute brightness by only considering gradients, precise spatial information by binning the gradients and precise values of the gradients by normalizing the histograms.
Convolutional neural networks (CNNs) trained in a supervised manner~\cite{LeCun_NC1989, Krizhevsky_NIPS2012} are expected to discard information irrelevant for the task they are solving~\cite{Zeiler_ECCV2014,Mahendran_CVPR2015,Springenberg_ICLR2015}.

In this paper we propose a new approach to analyze which information is preserved by a feature representation and which information is discarded.
We train neural networks to invert feature representations in the following sense.
Given a feature vector, the network is trained to predict the \emph{expected pre-image}, that is, the (weighted) average of all natural images which could have produced the given feature vector.
The content of this expected pre-image shows image properties which can be confidently inferred from the feature vector.
The amount of blur corresponds to the level of invariance of the feature representation.
We obtain further insights into the structure of the feature space, as we apply the networks to perturbed feature vectors, to interpolations between two feature vectors, or to random feature vectors.


\begin{figure}[]
\begin{center}
\setlength{\tabcolsep}{0.05cm}
\renewcommand{\arraystretch}{1}
\footnotesize{
  \begin{tabular}{cccc}
  HOG & SIFT & AlexNet-\conv3 & AlexNet-\fc8\\
  {\includegraphics[width=0.24\linewidth]{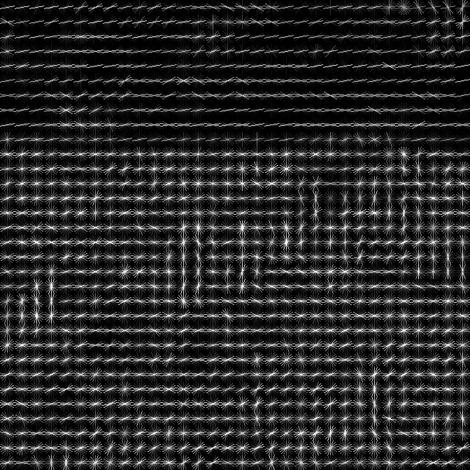}} &
  {\includegraphics[width=0.24\linewidth]{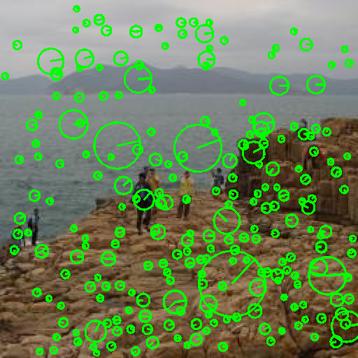}} &
  {\includegraphics[width=0.24\linewidth]{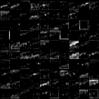}} &
  {\includegraphics[width=0.24\linewidth]{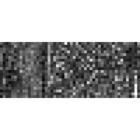}}
  \\
  {\includegraphics[width=0.24\linewidth]{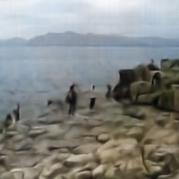}} &
  {\includegraphics[width=0.24\linewidth]{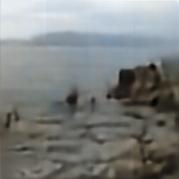}} &
  {\includegraphics[width=0.24\linewidth]{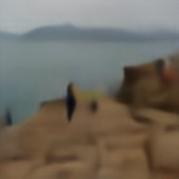}} &
  {\includegraphics[width=0.24\linewidth]{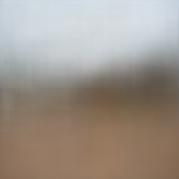}}\\\vspace{-6mm}
  \end{tabular}
}
  \caption{We train convolutional networks to reconstruct images from different feature representations. \textbf{Top row:} Input features. \textbf{Bottom row:} Reconstructed image.
  Reconstructions from HOG and SIFT are very realistic.
  Reconstructions from AlexNet preserve color and rough object positions even when reconstructing from higher layers.}
  \label{fig:teaser}
\end{center}
\vspace{-5mm}
\end{figure}


We apply our inversion method to AlexNet~\cite{Krizhevsky_NIPS2012}, a convolutional network trained for classification on ImageNet, as well as to three widely used computer vision features: histogram of oriented gradients (HOG)~\cite{DalalTriggs_2005, Felzenszwalb_PAMI2010}, scale invariant feature transform (SIFT)~\cite{Lowe_2004}, and local binary patterns (LBP)~\cite{Ojala_PAMI2002}.
The SIFT representation comes as a non-uniform, sparse set of oriented keypoints with their corresponding descriptors at various scales. This is an additional challenge for the inversion task.
LBP features are not differentiable with respect to the input image. Thus, existing methods based on gradients of representations~\cite{Mahendran_CVPR2015} could not be applied to them.


\subsection{Related work}

Our approach is related to a large body of work on inverting neural networks.
These include works making use of backpropagation or sampling~\cite{Lee_1994,Linden_1989,Lu_1999,Williams_1986,Jensen_1999,Varkonyi_2009} and, most similar to our approach, other neural networks~\cite{Bishop_1995}.
However, only recent advances in neural network architectures allow us to invert a modern large convolutional network with another network.

Our approach is not to be confused with the DeconvNet~\cite{Zeiler_ECCV2014}, which propagates high level activations backward through a network to identify parts of the image responsible for the activation.
In addition to the high-level feature activations, this reconstruction process uses extra information about maxima locations in intermediate max-pooling layers.
This information has been shown to be crucial for the approach to work ~\cite{Springenberg_ICLR2015}.
A visualization method similar to DeconvNet is by Springenberg et al.~\cite{Springenberg_ICLR2015}, yet it also makes use of intermediate layer activations.

Mahendran and Vedaldi~\cite{Mahendran_CVPR2015} invert a differentiable image representation $\Phi$ using gradient descent.
Given a feature vector $\Phi_0$, they seek for an image $x^*$ which minimizes a loss function~-- the squared Euclidean distance between $\Phi_0$ and $\Phi(x)$ plus a regularizer enforcing a natural image prior.
This method is fundamentally different from our approach in that it optimizes the difference between the feature vectors, not the image reconstruction error.
Additionally, it includes a hand-designed natural image prior, while in our case the network implicitly learns such a prior.
Technically, it involves optimization at test time, which requires computing the gradient of the feature representation and makes it relatively slow (the authors report 6s per image on a GPU).
In contrast, the presented approach is only costly when training the inversion network.
Reconstruction from a given feature vector just requires a single forward pass through the network, which takes roughly $5$ms per image on a GPU.
The method of~\cite{Mahendran_CVPR2015} requires gradients of the feature representation, therefore it could not be directly applied to non-differentiable representations such as LBP, or recordings from a real brain~\cite{Nishimoto_2011}.

There has been research on inverting various traditional computer vision representations: HOG and dense SIFT~\cite{Vondrick_ECCV2013}, keypoint-based SIFT~\cite{Weinzaepfel_CVPR2011}, Local Binary Descriptors~\cite{dAngelo_PAMI2014}, Bag-of-Visual-Words~\cite{Kato_CVPR2014}.
All these methods are either tailored for inverting a specific feature representation or restricted to shallow representations, while our method can be applied to any feature representation.


\section{Method}
Denote by $(\img,\, \feat)$ random variables representing a natural image and its feature vector, and denote their joint probability distribution by $p(\img, \feat) = p(\img)p(\feat|\img)$.
Here $p(\img)$ is the distribution of natural images and $p(\feat|\img)$ is the distribution of feature vectors given an image.
As a special case, $\feat$ may be a deterministic function of $\img$.
Ideally we would like to find $p(\img | \feat)$, but direct application of Bayes' theorem is not feasible.
Therefore in this paper we resort to a point estimate $\fnet(\feat)$ which minimizes the following mean squared error objective:
\begin{equation}
 \ex_{\img, \feat}\, ||\img - f(\feat)||^2
 \label{eq:ideal_loss}
\end{equation}
The minimizer of this loss is the conditional expectation:
\begin{equation}
 \hat{\fnet}(\feat_0) = \ex_\img\, [ \img\, |\, \feat = \feat_0],
\end{equation}
that is, the \emph{expected pre-image}.

Given a training set of images and their features $\{\img_i,\, \feat_i\}$, we learn the weights $\weights$ of an an up-convolutional network $\fnet(\feat, \weights)$ to minimize a Monte-Carlo estimate of the loss~\eqref{eq:ideal_loss}:
\begin{equation}
 \hat{\weights} = \arg\min\limits_\weights\; \sum\limits_i ||\img_i - \fnet(\feat_i, \weights)||_2^2.
 \label{eq:loss}
\end{equation}
This means that simply training the network to predict images from their feature vectors results in estimating the expected pre-image.

\subsection{Feature representations to invert}


\textbf{Shallow features.}
We invert three traditional computer vision feature representations: histogram of oriented gradients (HOG), scale invariant feature transform (SIFT), and local binary patterns (LBP).
We chose these features for a reason. There has been work on inverting HOG, so we can compare to existing approaches. LBP is interesting because it is not differentiable, and hence gradient-based methods cannot invert it. SIFT is a keypoint-based representation, so the network has to stitch different keypoints into a single smooth image.

For all three methods we use implementations from the \emph{VLFeat} library~\cite{VLfeat} with the default settings.
More precisely, we use the HOG version from Felzenszwalb et al.~\cite{Felzenszwalb_PAMI2010} with cell size $8$, the version of SIFT which is very similar to the original implementation of Lowe~\cite{Lowe_2004} and the LBP version similar to Ojala et al.~\cite{Ojala_PAMI2002} with cell size $16$.
Before extracting the features we convert images to grayscale.
More details can be found in the supplementary material.

\textbf{AlexNet.}
We also invert the representation of the AlexNet network~\cite{Krizhevsky_NIPS2012} trained on ImageNet, available at the \emph{Caffe}~\cite{caffe} website.\,\footnote{More precisely, we used CaffeNet, which is almost identical to the original AlexNet.}
It consists of $5$ convolutional layers and $3$ fully connected layers, with rectified linear units (ReLUs) after each layer, and local contrast normalization or max-pooling after some of them.
Exact architecture is shown in the supplementary material.
In what follows, when we say `output of the layer', we mean the output of the last processing step of this layer.
For example, the output of the first convolutional layer \conv1 would be the result after ReLU, pooling and normalization, and the output of the first fully connected layer \fc6 is after ReLU.
\fc8 denotes the last layer, before the softmax.

\subsection{Network architectures and training}

An up-convolutional layer, also often referred to as `deconvolutional', is a combination of upsampling and convolution~\cite{Dosovitskiy_CVPR2015}.
We upsample a feature map by a factor $2$ by replacing each value by a $2 \times 2$ block with the original value in the top left corner and all other entries equal to zero.
Architecture of one of our up-convolutional networks is shown in Table~\ref{tbl:fc8_arch}.
Architectures of other networks are shown in the supplementary material.

\textbf{HOG and LBP.}
For an image of size $W \times H$, HOG and LBP features of an image form 3-dimensional arrays of sizes $\ceil{W/8} \times \ceil{H/8} \times 31$ and $\ceil{W/16} \times \ceil{H/16} \times 58$, respectively.
We use similar CNN architectures for inverting both feature representations.
The networks include a contracting part, which processes the input features through a series of convolutional layers with occasional stride of $2$, resulting in a feature map $64$ times smaller than the input image.
Then the expanding part of the network again up-samples the feature map to the full image resolution by a series of up-convolutional layers.
The contracting part allows the network to aggregate information over large regions of the input image.
We found this is necessary to successfully estimate the absolute brightness.

\textbf{Sparse SIFT.}
Running the SIFT detector and descriptor on an image gives a set of $N$ keypoints, where the $i$-th keypoint is described by its coordinates $(x_i, y_i)$, scale $s_i$, orientation $\alpha_i$, and a feature descriptor $\siftfeat_i$ of dimensionality $D$.
In order to apply a convolutional network, we arrange the keypoints on a grid.
We split the image into cells of size $\ds \times \ds$ (we used $\ds = 4$ in our experiments), this yields $\ceil{W/\ds} \times \ceil{H/\ds}$ cells.
In the rare cases when there are several keypoints in a cell, we randomly select one.
We then assign a vector to each of the cells: a zero vector to a cell without a keypoint and a vector $(\siftfeat_i, x_i \mod \ds, y_i \mod \ds, \sin \alpha_i, \cos \alpha_i, \log s_i)$ to a cell with a keypoint.
This results in a feature map $\siftmap$ of size $\ceil{W/\ds} \times \ceil{H/\ds} \times (D + 5)$.
Then we apply a CNN to $\siftmap$, as described above.

\begin{table}
   \begin{center}
   \setlength{\tabcolsep}{0.15cm}
  \resizebox{0.98\linewidth}{!}{%
  \begin{tabular}{|l|cc|cc|c|}
      \hline
      Layer      & Input               & InSize               & K    & S    & OutSize              \\
      \hline                           
      fc1        & AlexNet-\fc8        & $1000$               & $-$  & $-$ & $4096$                \\
      fc2        & fc1                 & $4096$               & $-$  & $-$ & $4096$                \\
      fc3        & fc2                 & $4096$               & $-$  & $-$ & $4096$                \\
      reshape    & fc3                 & $4096$               & $-$  & $-$ & $4 \st 4 \st 256$     \\
      \hline                           
      upconv1    & reshape             & $4 \st 4 \st 256$    & $5$  & $2$ & $8 \st 8 \st 256$     \\
      upconv2    & upconv1             & $8 \st 8 \st 256$    & $5$  & $2$ & $16 \st 16 \st 128$   \\
      upconv3    & upconv2             & $16 \st 16 \st 128$  & $5$  & $2$ & $32 \st 32 \st 64$    \\
      upconv4    & upconv3             & $32 \st 32 \st 64$   & $5$  & $2$ & $64 \st 64 \st 32$    \\
      upconv5    & upconv4             & $64 \st 64 \st 32$   & $5$  & $2$ & $128 \st 128 \st 3$   \\
      \hline
    \end{tabular}}
  \end{center}
  \caption{Network for reconstructing from AlexNet \fc8 features. K stands for kernel size, S for stride.}
  \label{tbl:fc8_arch}
\end{table}

\textbf{AlexNet.}
To reconstruct from each layer of AlexNet we trained a separate network.
We used two basic architectures: one for reconstructing from convolutional layers and one for reconstructing from fully connected layers.
The network for reconstructing from fully connected layers contains three fully connected layers and $5$ up-convolutional layers, as shown in Table~\ref{tbl:fc8_arch}.
The network for reconstructing from convolutional layers consists of three convolutional and several up-convolutional layers (the exact number depends on the layer to reconstruct from).
Filters in all (up-)convolutional layers have $5 \times 5$ spatial size.
After each layer we apply leaky ReLU nonlinearity with slope $0.2$, that is, $r(x) = x$ if $x \geqslant 0$ and $r(x) = 0.2 \cdot x$ if $x<0$.

\textbf{Training details.}
We trained networks using a modified version of Caffe~\cite{caffe}.
As training data we used the ImageNet~\cite{imagenet} training set.
In some cases we predicted downsampled images to speed up computations.
We used the Adam~\cite{Kingma_ICLR2015} optimizer with $\beta_1 = 0.9$, $\beta_2 = 0.999$ and mini-batch size $64$.
For most networks we found an initial learning rate $\lambda = 0.001$ to work well.
We gradually decreased the learning rate towards the end of training.
The duration of training depended on the network: from $15$ epochs (passes through the dataset) for shallower networks to $60$ epochs for deeper ones.

\textbf{Quantitative evaluation.}
As a quantitative measure of performance we used the average normalized reconstruction error, that is the mean of $||x_i - f(\Phi(x_i))||_2 / N$, where $x_i$ is an example from the test set, $f$ is the function implemented by the inversion network and $N$ is a normalization coefficient equal to the average Euclidean distance between images in the test set.
The test set we used for quantitative and qualitative evaluations is a subset of the ImageNet validation set.

\begin{table}[b]
\begin{center}
\vspace{-0.3cm}
\begin{tabular}{|c|c||c|c|c|}
\hline
\footnotesize{Hoggles~\cite{Vondrick_ECCV2013}} & \footnotesize{$HOG^{-1}$~\cite{Mahendran_CVPR2015}} & \footnotesize{HOG our} & \footnotesize{SIFT our} & \footnotesize{LBP our} \\
\hline
$0.61$ & $0.63$ & $0.24$ & $0.28$ & $0.38$ \\
\hline
\end{tabular}
\end{center}
\vspace*{-0.3cm}
\caption{Normalized error of different methods when reconstructing from HOG.}
\label{tbl:shallow}
\end{table}

\section{Experiments: shallow representations}
Figures~\ref{fig:teaser} and~\ref{fig:invert_shallow} show reconstructions of several images from the ImageNet validation set.
Normalized reconstruction error of different approaches is shown in Table~\ref{tbl:shallow}.
Clearly, our method significantly outperforms existing approaches.
This is to be expected, since our method explicitly aims to minimize the reconstruction error.

\begin{figure*}[t]
\begin{center}
\setlength{\tabcolsep}{0.03cm}
\renewcommand{\arraystretch}{0.5}
  \begin{tabular}{ccccc}
  Image & HOG & Hoggles~\cite{Vondrick_ECCV2013} & $HOG^{-1}$~\cite{Mahendran_CVPR2015} & Our
  \\[1mm]
  {\includegraphics[width=0.19\linewidth]{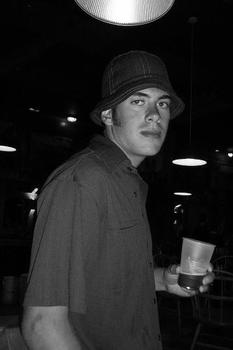}} &
  {\includegraphics[width=0.19\linewidth]{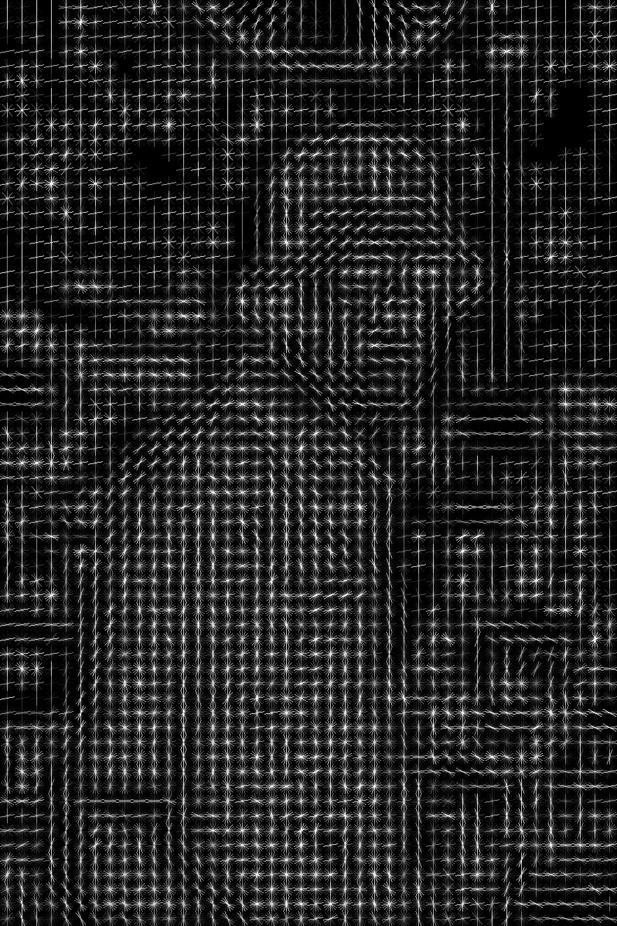}} &
  {\includegraphics[width=0.19\linewidth]{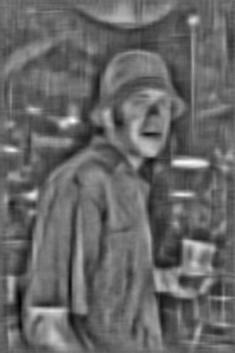}} &
  {\includegraphics[width=0.19\linewidth]{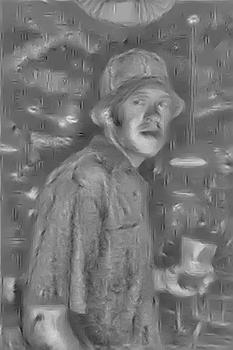}} &
  {\includegraphics[width=0.19\linewidth]{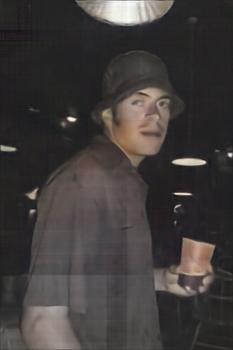}}
  \end{tabular}
\end{center}
\vspace*{-0.5cm}
\caption{Reconstructing an image from its HOG descriptors with different methods.}
\label{fig:hog_guy}
\end{figure*}

\textbf{Colorization.}
As mentioned above, we compute the features based on grayscale images, but the task of the networks is to reconstruct the color images.
The features do not contain any color information, so to predict colors the network has to analyze the content of the image and make use of a natural image prior it learned during training.
It does successfully learn to do so, as can be seen in Figures~\ref{fig:teaser} and~\ref{fig:invert_shallow}.
Quite often the colors are predicted correctly, especially for sky, sea, grass, trees.
In other cases, the network cannot predict the color (for example, people in the top row of Figure~\ref{fig:invert_shallow}) and leaves some areas gray.
Occasionally the network predicts the wrong color, such as in the bottom row of Figure~\ref{fig:invert_shallow}.

\textbf{HOG.}
Figure~\ref{fig:hog_guy} shows an example image, its HOG representation, the results of inversion with existing methods~\cite{Vondrick_ECCV2013, Mahendran_CVPR2015} and with our approach.
Most interestingly, the network is able to reconstruct the overall brightness of the image very well, for example the dark regions are reconstructed dark.
This is quite surprising, since the HOG descriptors are normalized and should not contain information about absolute brightness.

Normalization is always performed with a smoothing 'epsilon', so one might imagine that some information about the brightness is present even in the normalized features.
We checked that the network does not make use of this information: multiplying the input image by $10$ or $0.1$ hardly changes the reconstruction.
Therefore, we  hypothesize that the network reconstructs the overall brightness by 1) analyzing the distribution of the HOG features (if in a cell there is similar amount of gradient in all directions, it is probably noise; if there is one dominating gradient, it must actually be in the image), 2) accumulating gradients over space: if there is much black-to-white gradient in one direction, then probably the brightness in that direction goes from dark to bright and 3) using semantic information.

\begin{figure}[]
\begin{center}
\setlength{\tabcolsep}{0.03cm}
\renewcommand{\arraystretch}{0.5}
  \begin{tabular}{cccc}
  Image & HOG our & SIFT our & LBP our
  \\
  {\includegraphics[width=0.242\linewidth]{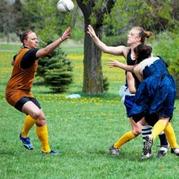}} &
  {\includegraphics[width=0.242\linewidth]{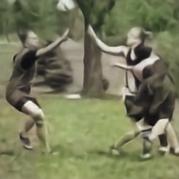}} &
  {\includegraphics[width=0.242\linewidth]{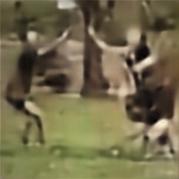}} &
  {\includegraphics[width=0.242\linewidth]{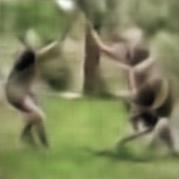}}
  \\
  {\includegraphics[width=0.242\linewidth]{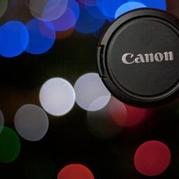}} &
  {\includegraphics[width=0.242\linewidth]{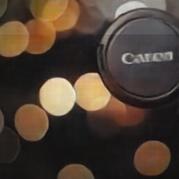}} &
  {\includegraphics[width=0.242\linewidth]{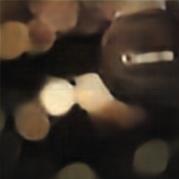}} &
  {\includegraphics[width=0.242\linewidth]{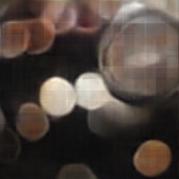}}
  \end{tabular}
\end{center}
\vspace*{-0.5cm}
\caption{Inversion of shallow image representations. Note how in the first row the color of grass and trees is predicted correctly in all cases, although it is not contained in the features.}
\label{fig:invert_shallow}
\end{figure}

\begin{figure}[]
\begin{center}
\setlength{\tabcolsep}{0.03cm}
\renewcommand{\arraystretch}{0.5}
  \begin{tabular}{ccccc}
  \\
  {\includegraphics[width=0.49\linewidth]{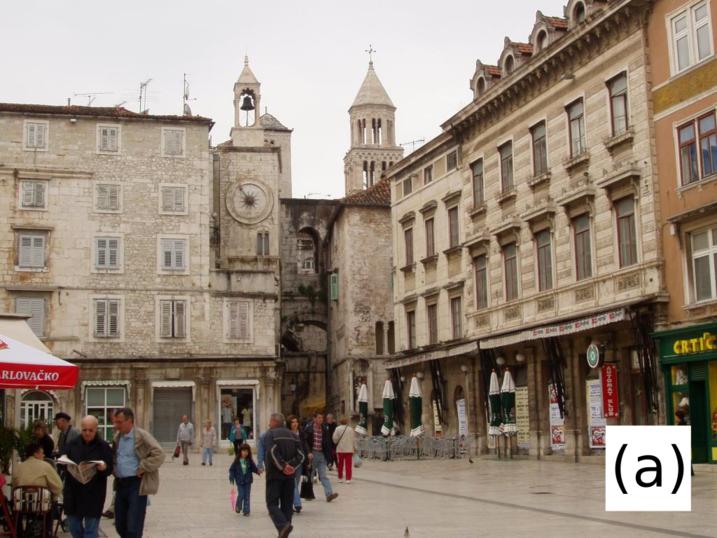}} &
  {\includegraphics[width=0.49\linewidth]{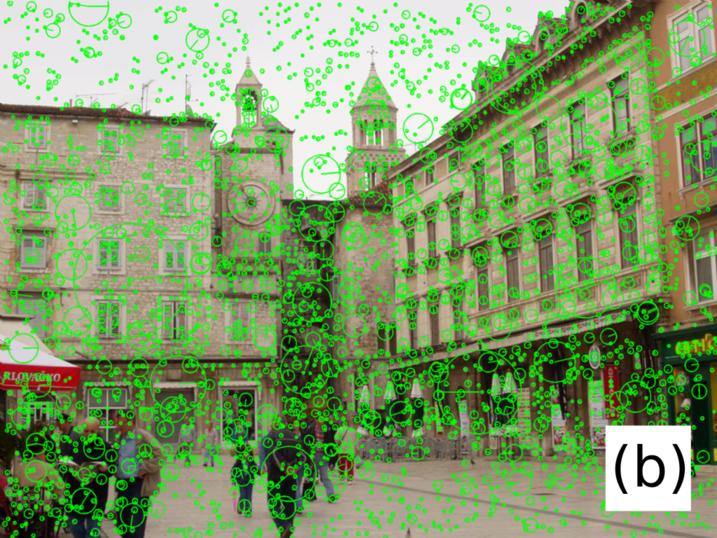}}
  \\
  {\includegraphics[width=0.49\linewidth]{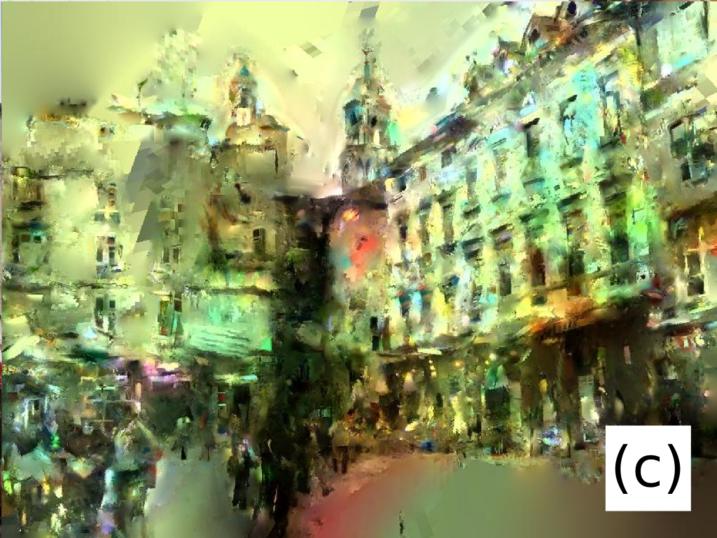}} &
  {\includegraphics[width=0.49\linewidth]{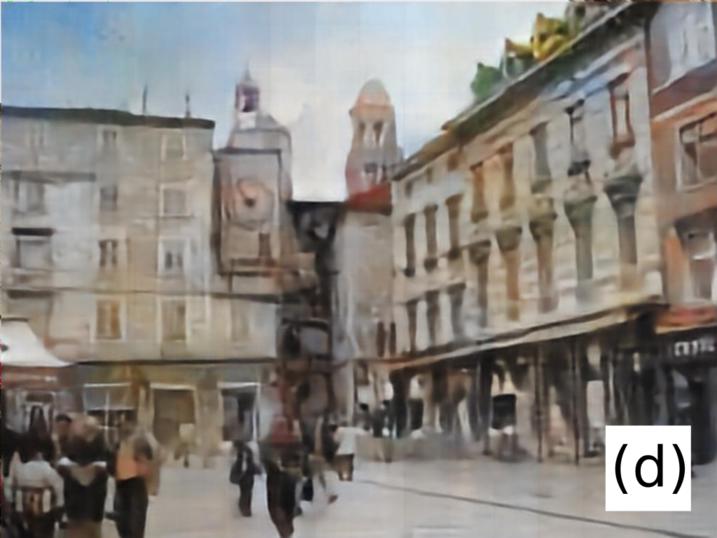}}
  \end{tabular}
\end{center}
\vspace*{-0.5cm}
\caption{Reconstructing an image from SIFT descriptors with different methods. \textbf{(a)} an image, \textbf{(b)} SIFT keypoints, \textbf{(c)} reconstruction of~\cite{Weinzaepfel_CVPR2011}, \textbf{(d)} our reconstruction.}
\label{fig:sift_qualitative}
\end{figure}

\begin{figure*}[tb] 
\begin{center}
\setlength{\tabcolsep}{0.04cm}
\renewcommand{\arraystretch}{0.55}
  \begin{centering}
  \begin{tabular}{ccccccccc}
  Image & \conv1 & \conv2 & \conv3 & \conv4 & \conv5 & \fc6 & \fc7 & \fc8
  \\
  {\includegraphics[width=0.095\linewidth]{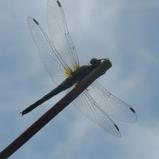}} &
  {\includegraphics[width=0.095\linewidth]{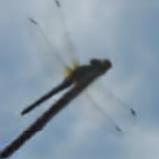}} &
  {\includegraphics[width=0.095\linewidth]{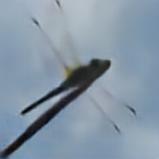}} &
  {\includegraphics[width=0.095\linewidth]{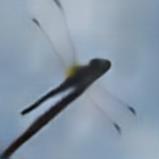}} &
  {\includegraphics[width=0.095\linewidth]{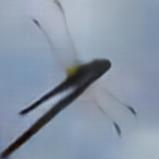}} &
  {\includegraphics[width=0.095\linewidth]{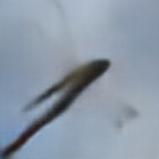}} &
  {\includegraphics[width=0.095\linewidth]{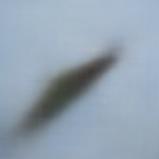}} &
  {\includegraphics[width=0.095\linewidth]{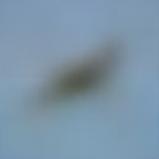}} &
  {\includegraphics[width=0.095\linewidth]{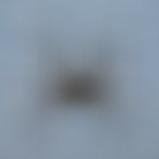}}
    \\
  {\includegraphics[width=0.095\linewidth]{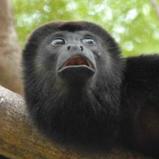}} &
  {\includegraphics[width=0.095\linewidth]{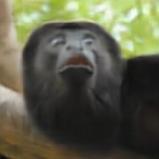}} &
  {\includegraphics[width=0.095\linewidth]{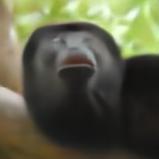}} &
  {\includegraphics[width=0.095\linewidth]{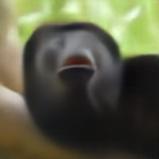}} &
  {\includegraphics[width=0.095\linewidth]{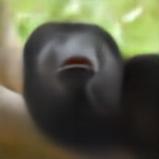}} &
  {\includegraphics[width=0.095\linewidth]{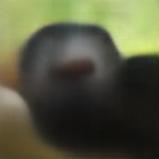}} &
  {\includegraphics[width=0.095\linewidth]{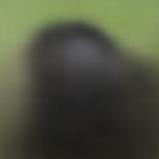}} &
  {\includegraphics[width=0.095\linewidth]{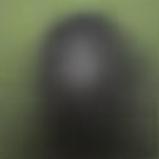}} &
  {\includegraphics[width=0.095\linewidth]{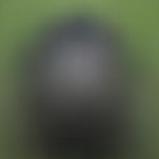}}
  \\
  {\includegraphics[width=0.095\linewidth]{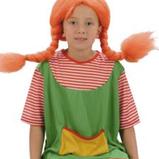}} &
  {\includegraphics[width=0.095\linewidth]{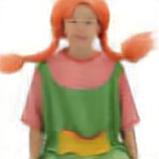}} &
  {\includegraphics[width=0.095\linewidth]{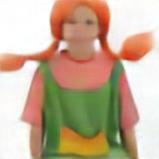}} &
  {\includegraphics[width=0.095\linewidth]{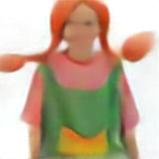}} &
  {\includegraphics[width=0.095\linewidth]{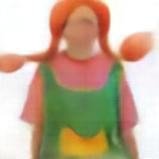}} &
  {\includegraphics[width=0.095\linewidth]{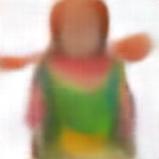}} &
  {\includegraphics[width=0.095\linewidth]{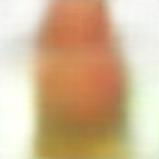}} &
  {\includegraphics[width=0.095\linewidth]{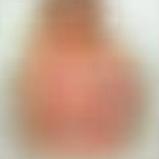}} &
  {\includegraphics[width=0.095\linewidth]{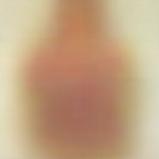}}
  \\
  {\includegraphics[width=0.095\linewidth]{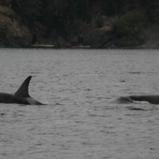}} &
  {\includegraphics[width=0.095\linewidth]{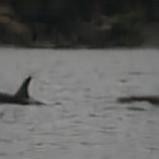}} &
  {\includegraphics[width=0.095\linewidth]{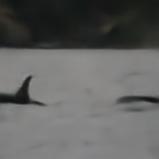}} &
  {\includegraphics[width=0.095\linewidth]{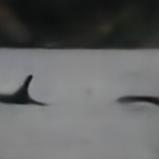}} &
  {\includegraphics[width=0.095\linewidth]{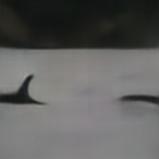}} &
  {\includegraphics[width=0.095\linewidth]{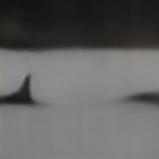}} &
  {\includegraphics[width=0.095\linewidth]{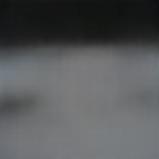}} &
  {\includegraphics[width=0.095\linewidth]{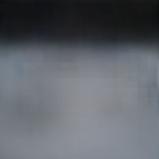}} &
  {\includegraphics[width=0.095\linewidth]{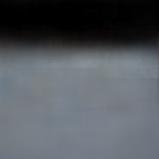}}\vspace{-0.5cm}
  \end{tabular}
  \end{centering}
\end{center}
   \caption{Reconstructions from different layers of AlexNet.}
\label{fig:recon_examples}
\end{figure*}

\begin{figure*}[tb] 
\begin{center}
\setlength{\tabcolsep}{0.03cm}
\renewcommand{\arraystretch}{0.5}
  \begin{tabular}{cccccccccc}
     & Image & \conv1 & \conv2 & \conv3 & \conv4 & \conv5 & \fc6 & \fc7 & \fc8
        \\
  Our &
  \raisebox{-.5\height}{\includegraphics[width=0.095\linewidth]{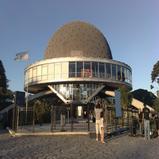}} &
  \raisebox{-.5\height}{\includegraphics[width=0.095\linewidth]{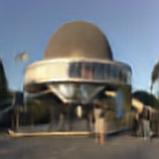}} &
  \raisebox{-.5\height}{\includegraphics[width=0.095\linewidth]{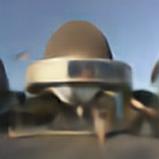}} &
  \raisebox{-.5\height}{\includegraphics[width=0.095\linewidth]{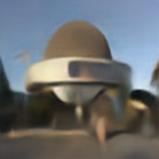}} &
  \raisebox{-.5\height}{\includegraphics[width=0.095\linewidth]{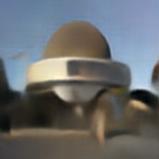}} &
  \raisebox{-.5\height}{\includegraphics[width=0.095\linewidth]{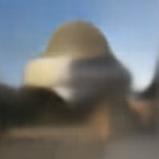}} &
  \raisebox{-.5\height}{\includegraphics[width=0.095\linewidth]{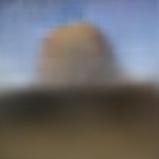}} &
  \raisebox{-.5\height}{\includegraphics[width=0.095\linewidth]{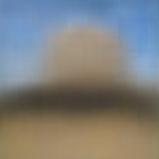}} &
  \raisebox{-.5\height}{\includegraphics[width=0.095\linewidth]{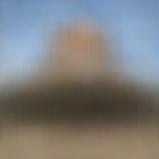}} \vspace*{0.095cm}
  \\
  \cite{Mahendran_CVPR2015} &
  \raisebox{-.5\height}{\includegraphics[width=0.095\linewidth]{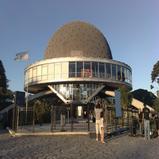}} &
  \raisebox{-.5\height}{\includegraphics[width=0.095\linewidth]{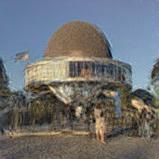}} &
  \raisebox{-.5\height}{\includegraphics[width=0.095\linewidth]{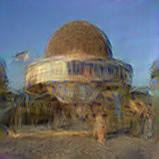}} &
  \raisebox{-.5\height}{\includegraphics[width=0.095\linewidth]{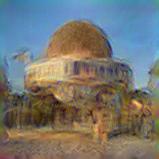}} &
  \raisebox{-.5\height}{\includegraphics[width=0.095\linewidth]{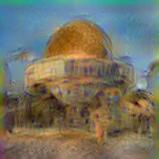}} &
  \raisebox{-.5\height}{\includegraphics[width=0.095\linewidth]{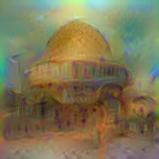}} &
  \raisebox{-.5\height}{\includegraphics[width=0.095\linewidth]{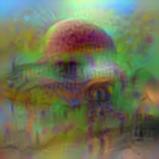}} &
  \raisebox{-.5\height}{\includegraphics[width=0.095\linewidth]{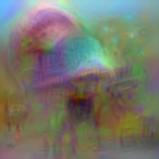}} &
  \raisebox{-.5\height}{\includegraphics[width=0.095\linewidth]{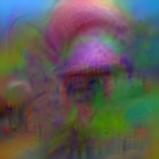}} \vspace*{0.095cm}
   \\
  AE &
  \raisebox{-.5\height}{\includegraphics[width=0.095\linewidth]{compressed_resources/images/1049_0orig.jpg}} &
  \raisebox{-.5\height}{\includegraphics[width=0.095\linewidth]{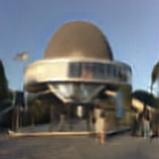}} &
  \raisebox{-.5\height}{\includegraphics[width=0.095\linewidth]{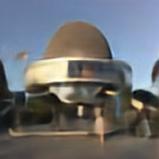}} &
  \raisebox{-.5\height}{\includegraphics[width=0.095\linewidth]{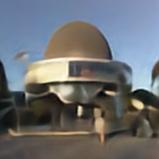}} &
  \raisebox{-.5\height}{\includegraphics[width=0.095\linewidth]{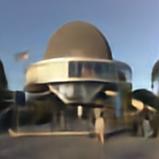}} &
  \raisebox{-.5\height}{\includegraphics[width=0.095\linewidth]{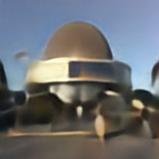}} &
  \raisebox{-.5\height}{\includegraphics[width=0.095\linewidth]{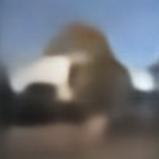}} &
  \raisebox{-.5\height}{\includegraphics[width=0.095\linewidth]{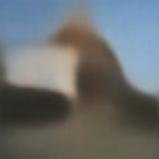}} &
  \raisebox{-.5\height}{\includegraphics[width=0.095\linewidth]{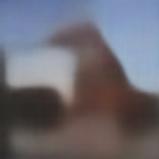}} \vspace*{-0.5cm}
   \end{tabular}
\end{center}
   \caption{Reconstructions from layers of AlexNet with our method (top), ~\cite{Mahendran_CVPR2015} (middle), and autoencoders (bottom).}
\label{fig:recon_compare}
\end{figure*}

\textbf{SIFT.}
Figure~\ref{fig:sift_qualitative} shows an image, the detected SIFT keypoints and the resulting reconstruction.
There are roughly $3000$ keypoints detected in this image.
Although made from a sparse set of keypoints, the reconstruction looks very natural, just a little blurry.
To achieve such a clear reconstruction the network has to properly rotate and scale the descriptors and then stitch them together.
Obviously it successfully learns to do this.

For reference we also show a result of another existing method~\cite{Weinzaepfel_CVPR2011} for reconstructing images from sparse SIFT descriptors.
The results are not directly comparable: while we use the SIFT detector providing circular keypoints, Weinzaepfel et al.~\cite{Weinzaepfel_CVPR2011} use the Harris affine keypoint detector which yields elliptic keypoints, and the number and the locations of the keypoints may be different from our case.
However, the rough number of keypoints is the same, so a qualitative comparison is still valid.

\section{Experiments: AlexNet}
\label{sec:experiments_alexnet}
We applied our inversion method to different layers of AlexNet and performed several additional experiments to better understand the feature representations.
More results are shown in the supplementary material.

\subsection{Reconstructions from different layers}
Figure~\ref{fig:recon_examples} shows reconstructions from various layers of AlexNet.
When using features from convolutional layers, the reconstructed images look very similar to the input, but lose fine details as we progress to higher layers.
There is an obvious drop in reconstruction quality when going from \conv5 to \fc6.
However, the reconstructions from higher convolutional layers and even fully connected layers preserve color and the approximate object location very well.
Reconstructions from \fc7 and \fc8 still look similar to the input images, but blurry.
This means that high level features are much less invariant to color and pose than one might expect: in principle fully connected layers need not preserve any information about colors and locations of objects in the input image.
This is somewhat in contrast with the results of~\cite{Mahendran_CVPR2015}, as shown in Figure~\ref{fig:recon_compare}.
While their reconstructions are sharper, the color and position are completely lost in reconstructions from higher layers.

\begin{figure}[tb] 
\centering
\includegraphics[width=0.95\linewidth]{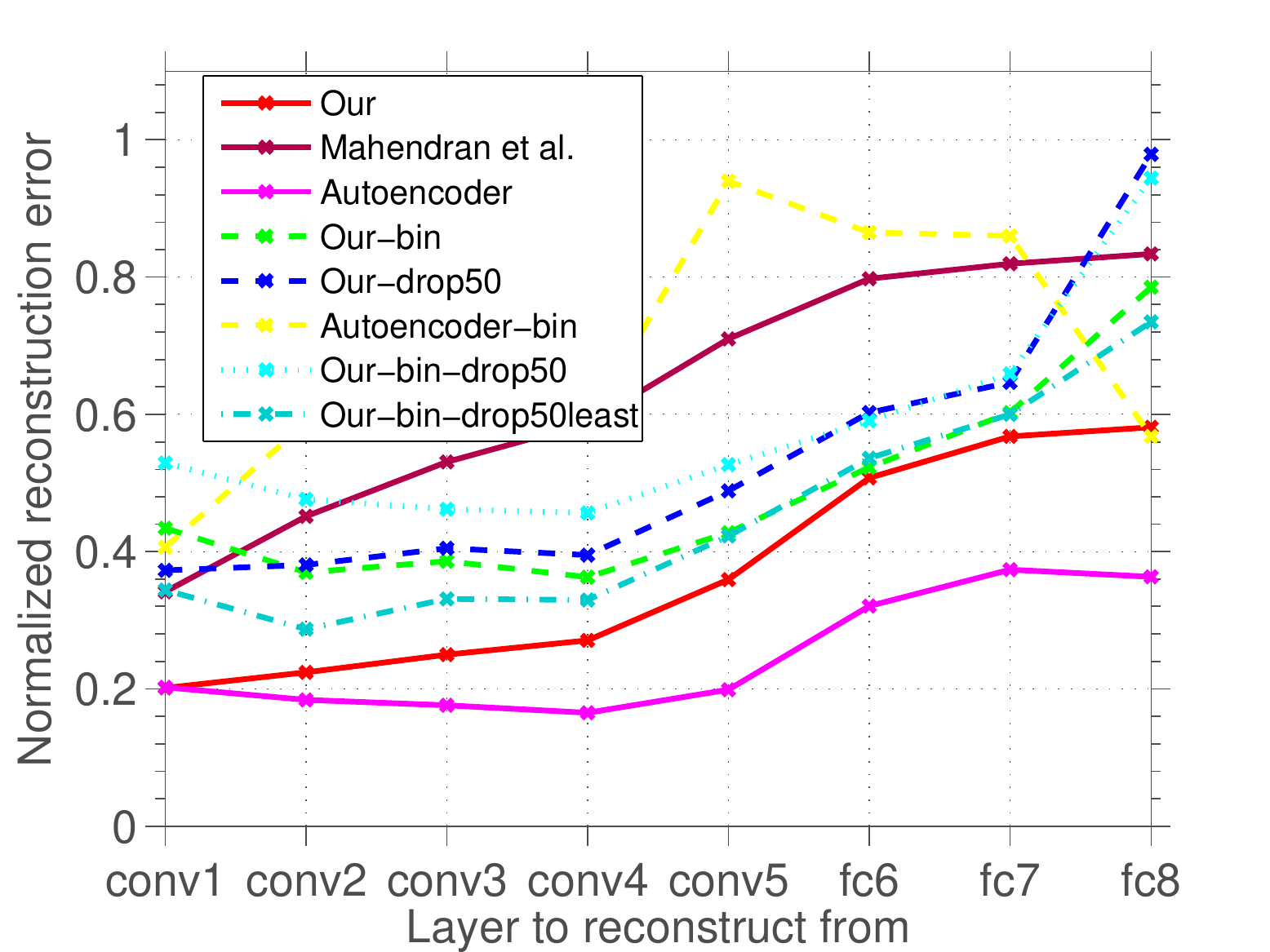}
\caption{Average normalized reconstruction error depending on the network layer.}
  \label{fig:recon_plots}
\end{figure}%

For quantitative evaluation before computing the error we up-sample reconstructions to input image size with bilinear interpolation.
Error curves shown in Figure~\ref{fig:recon_plots} support the conclusions made above.
When reconstructing from \fc6, the error is roughly twice as large as from \conv5.
Even when reconstructing from \fc8, the error is fairly low because the network manages to get the color and the rough placement of large objects in images right.
For lower layers, the reconstruction error of~\cite{Mahendran_CVPR2015} is still much higher than of our method, even though visually the images look somewhat sharper.
The reason is that in their reconstructions the color and the precise placement of small details do not perfectly match the input image, which results in a large overall error.


\subsection{Autoencoder training}
Our inversion network can be interpreted as the decoder of the representation encoded by AlexNet.
The difference to an autoencoder is that the encoder part stays fixed and only the decoder is optimized.
For comparison we also trained autoencoders with the same architecture as our reconstruction nets, i.e., we also allowed the training to fine-tune the parameters of the AlexNet part.
This provides an upper bound on the quality of reconstructions we might expect from the inversion networks (with fixed AlexNet).

As shown in Figure~\ref{fig:recon_plots}, autoencoder training yields much lower reconstruction errors when reconstructing from higher layers.
Also the qualitative results in Figure~\ref{fig:recon_compare} show much better reconstructions with autoencoders.
Even from \conv5 features, the input image can be reconstructed almost perfectly.
When reconstructing from fully connected layers, the autoencoder results get blurred, too, due to the compressed representation, but by far not as much as with the fixed AlexNet weights.
The gap between the autoencoder training and the training with fixed AlexNet gives an  estimate of the amount of image information lost due to the training objective of the AlexNet, which is not based on reconstruction quality.

An interesting observation with autoencoders is that the reconstruction error is quite high even when reconstructing from \conv1 features, and the best reconstructions were actually obtained from \conv4.
Our explanation is that the convolution with stride $4$ and consequent max-pooling in \conv1 loses much information about the image.
To decrease the reconstruction error, it is beneficial for the network to slightly blur the image instead of guessing the details.
When reconstructing from deeper layers, deeper networks can learn a better prior resulting in slightly sharper images and slightly lower reconstruction error.
For even deeper layers, the representation gets too compressed and the error increases again.
We observed (not shown in the paper) that without stride $4$ in the first layer, the reconstruction error of autoencoders got much lower.


\begin{figure}
\begin{center}
\setlength{\tabcolsep}{0.03cm}
\renewcommand{\arraystretch}{0.5}
  \begin{tabular}{cccc}
  Image & all & top5 & notop5
  \\
  {\includegraphics[width=0.16\linewidth]{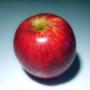}} &
  {\includegraphics[width=0.16\linewidth]{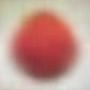}} &
  {\includegraphics[width=0.16\linewidth]{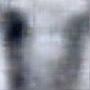}} &
  {\includegraphics[width=0.16\linewidth]{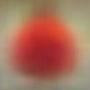}}
  \\
  \multicolumn{4}{c}{pomegranate (0.93)} \vspace*{0.2cm}
  \\
  {\includegraphics[width=0.16\linewidth]{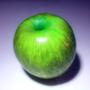}} &
  {\includegraphics[width=0.16\linewidth]{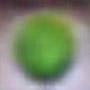}} &
  {\includegraphics[width=0.16\linewidth]{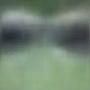}} &
  {\includegraphics[width=0.16\linewidth]{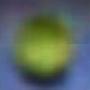}}
  \\
  \multicolumn{4}{c}{Granny Smith apple (0.99)} \vspace*{0.2cm}
  \\
  {\includegraphics[width=0.16\linewidth]{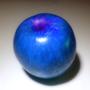}} &
  {\includegraphics[width=0.16\linewidth]{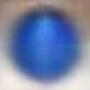}} &
  {\includegraphics[width=0.16\linewidth]{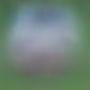}} &
  {\includegraphics[width=0.16\linewidth]{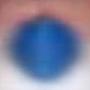}}
  \\
  \multicolumn{4}{c}{croquet ball (0.96)}\vspace{-5mm}
  \end{tabular}
\end{center}
  \caption{The effect of color on classification and reconstruction from layer \fc8. Left to right: input image, reconstruction from \fc8, reconstruction from $5$ largest activations in \fc8, reconstruction from all \fc8 activations except the $5$ largest ones. Below each row the network prediction and its confidence are shown.}
  \label{fig:apple}
\end{figure}

\begin{figure*} 
\scriptsize
\begin{center}
\setlength{\tabcolsep}{0.03cm}
\renewcommand{\arraystretch}{0.5}
  \begin{tabular}{cccccccccccccc}
     & Image & \conv3 & \conv4 & \conv5 & \fc6 & \fc7 & \fc8 & \conv3 & \conv4 & \conv5 & \fc6 & \fc7 & \fc8
  \\
  \shortstack{No \\ per- \\ turb} &
  \raisebox{-.2\height}{\includegraphics[width=0.067\linewidth]{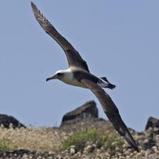}} &
  \;\;
  \raisebox{-.2\height}{\includegraphics[width=0.067\linewidth]{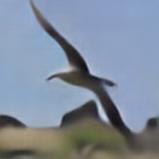}} &
  \raisebox{-.2\height}{\includegraphics[width=0.067\linewidth]{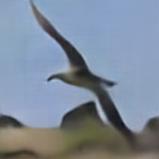}} &
  \raisebox{-.2\height}{\includegraphics[width=0.067\linewidth]{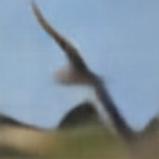}} &
  \raisebox{-.2\height}{\includegraphics[width=0.067\linewidth]{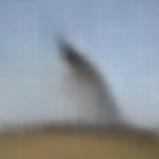}} &
  \raisebox{-.2\height}{\includegraphics[width=0.067\linewidth]{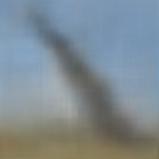}} &
  \raisebox{-.2\height}{\includegraphics[width=0.067\linewidth]{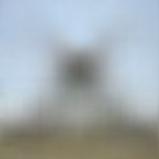}} &
  \;\;
  \raisebox{-.2\height}{\includegraphics[width=0.067\linewidth]{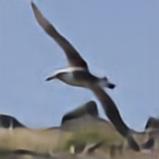}} &
  \raisebox{-.2\height}{\includegraphics[width=0.067\linewidth]{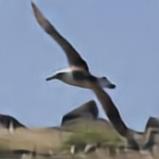}} &
  \raisebox{-.2\height}{\includegraphics[width=0.067\linewidth]{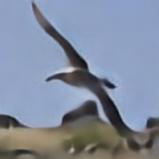}} &
  \raisebox{-.2\height}{\includegraphics[width=0.067\linewidth]{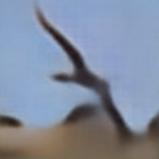}} &
  \raisebox{-.2\height}{\includegraphics[width=0.067\linewidth]{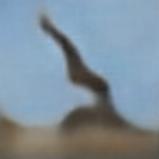}} &
  \raisebox{-.2\height}{\includegraphics[width=0.067\linewidth]{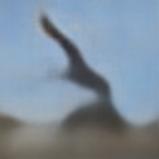}} \vspace*{0.06cm}
  \\
  Bin &
  \raisebox{-.5\height}{\includegraphics[width=0.067\linewidth]{compressed_resources/images/6_0orig.jpg}} &
  \;\;
  \raisebox{-.5\height}{\includegraphics[width=0.067\linewidth]{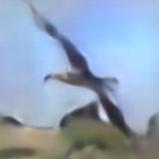}} &
  \raisebox{-.5\height}{\includegraphics[width=0.067\linewidth]{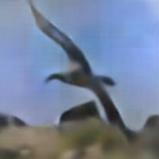}} &
  \raisebox{-.5\height}{\includegraphics[width=0.067\linewidth]{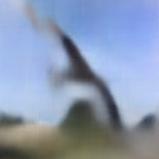}} &
  \raisebox{-.5\height}{\includegraphics[width=0.067\linewidth]{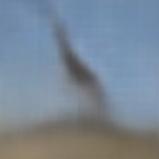}} &
  \raisebox{-.5\height}{\includegraphics[width=0.067\linewidth]{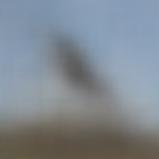}} &
  \raisebox{-.5\height}{\includegraphics[width=0.067\linewidth]{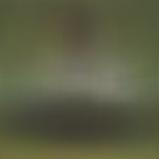}} &
  \;\;
  \raisebox{-.5\height}{\includegraphics[width=0.067\linewidth]{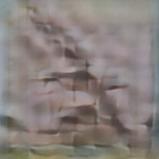}} &
  \raisebox{-.5\height}{\includegraphics[width=0.067\linewidth]{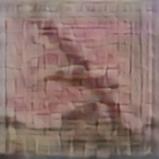}} &
  \raisebox{-.5\height}{\includegraphics[width=0.067\linewidth]{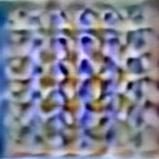}} &
  \raisebox{-.5\height}{\includegraphics[width=0.067\linewidth]{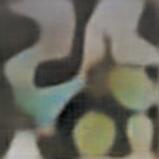}} &
  \raisebox{-.5\height}{\includegraphics[width=0.067\linewidth]{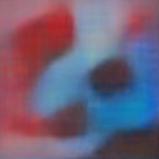}} &
  \raisebox{-.5\height}{\includegraphics[width=0.067\linewidth]{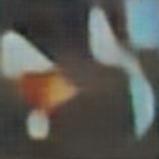}} \vspace*{0.06cm}
  \\
  \shortstack{Drop \\ 50} &
  \raisebox{-.3\height}{\includegraphics[width=0.067\linewidth]{compressed_resources/images/6_0orig.jpg}} &
  \;\;
  \raisebox{-.3\height}{\includegraphics[width=0.067\linewidth]{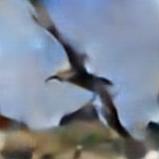}} &
  \raisebox{-.3\height}{\includegraphics[width=0.067\linewidth]{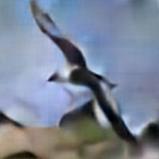}} &
  \raisebox{-.3\height}{\includegraphics[width=0.067\linewidth]{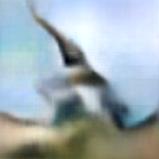}} &
  \raisebox{-.3\height}{\includegraphics[width=0.067\linewidth]{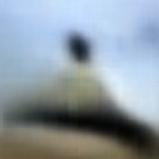}} &
  \raisebox{-.3\height}{\includegraphics[width=0.067\linewidth]{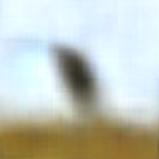}} &
  \raisebox{-.3\height}{\includegraphics[width=0.067\linewidth]{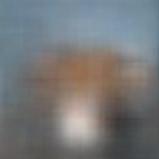}} &
  \;\;
  \raisebox{-.3\height}{\includegraphics[width=0.067\linewidth]{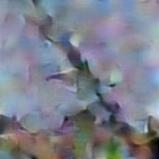}} &
  \raisebox{-.3\height}{\includegraphics[width=0.067\linewidth]{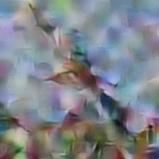}} &
  \raisebox{-.3\height}{\includegraphics[width=0.067\linewidth]{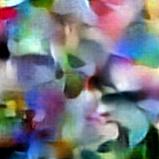}} &
  \raisebox{-.3\height}{\includegraphics[width=0.067\linewidth]{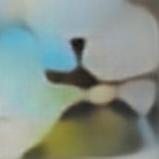}} &
  \raisebox{-.3\height}{\includegraphics[width=0.067\linewidth]{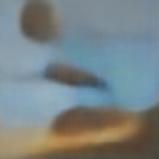}} &
  \raisebox{-.3\height}{\includegraphics[width=0.067\linewidth]{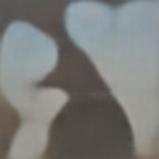}} \vspace*{0.1cm}
  \\
  & &  \multicolumn{6}{c}{\small{Fixed AlexNet}} & \multicolumn{6}{c}{\small{Autoencoder}}\vspace{-5mm}
  \end{tabular}
\end{center}
   \caption{Reconstructions from different layers of AlexNet with disturbed features. }
\label{fig:recon_perturb}
\end{figure*}

\subsection{Case study: Colored apple}
\label{subsec:color_position}
We performed a simple experiment illustrating how the color information influences classification and how it is preserved in the high level features.
We took an image of a red apple (Figure~\ref{fig:apple} top left) from Flickr and modified its hue to make it green or blue.
Then we extracted AlexNet \fc8 features of the resulting images.
Remind that \fc8 is the last layer of the network, so the \fc8 features, after application of softmax, give the network's prediction of class probabilities.
The largest activation, hence, corresponds to the network's prediction of the image class.
To check how class-dependent the results of inversion are, we passed three versions of each feature vector through the inversion network: 1) just the vector itself, 2) all activations except the $5$ largest ones set to zero, 3) the $5$ largest activations set to zero.

This leads to several conclusions.
First, color clearly can be very important for classification, so the feature representation of the network has to be sensitive to it, at least in some cases.
Second, the color of the image can be precisely reconstructed even from \fc8 or, equivalently, from the predicted class probabilities.
Third, the reconstruction quality does not depend much on the top predictions of the network but rather on the small probabilities of all other classes.
This is consistent with the 'dark knowledge' idea of~\cite{Hinton_arxiv2015}: small probabilities of non-predicted classes carry more information than the prediction itself.
More examples of this are shown in the supplementary material.


\subsection{Robustness of the feature representation} \label{sec:robustness}
We have shown that high level feature maps preserve rich information about the image.
How is this information represented in the feature vector?
It is difficult to answer this question precisely, but we can gain some insight by perturbing the feature representations in certain ways and observing images reconstructed from these perturbed features.
If perturbing the features in a certain way does not change the reconstruction much, then the perturbed property is not important.
For example, if setting a non-zero feature to zero does not change the reconstruction, then this feature does not carry information useful for the reconstruction.

We applied binarization and dropout.
To binarize the feature vector, we kept the signs of all entries and set their absolute values to a fixed number, selected such that the Euclidean norm of the vector remained unchanged (we tried several other strategies, and this one led to the best result).
For all layers except \fc8, feature vector entries are non-negative, hence, binarization just sets all non-zero entries to a fixed positive value.
To perform dropout, we randomly set $50 \%$ of the feature vector entries to zero and then normalize the vector to keep its Euclidean norm unchanged (again, we found this normalization to work best).
Qualitative results of these perturbations of features in different layers of AlexNet are shown in Figure~\ref{fig:recon_perturb}.
Quantitative results are shown in Figure~\ref{fig:recon_plots}.
Surprisingly, dropout leads to larger decrease in reconstruction accuracy than binarization, even in the layers where it had been applied during training.
In layers \fc7 and especially \fc6, binarization hardly changes the reconstruction quality at all.
Although it is known that binarized ConvNet features perform well in classification~\cite{Agrawal_ECCV2014}, it comes as a surprise that for reconstructing the input image the exact values of the features are not important.
In \fc6 virtually all information about the image is contained in the binary code given by the pattern of non-zero activations.
Figures~\ref{fig:recon_plots} and~\ref{fig:recon_perturb} show that this binary code only emerges when training with the classification objective and dropout, while autoencoders are very sensitive to perturbations in the features.

To test the robustness of this binary code, we applied binarization and dropout together.
We tried dropping out $50 \%$ random activations or $50 \%$ least non-zero activations and then binarizing.
Dropping out the $50\%$ least activations reduces the error much less than dropping out $50\%$ random activations and is even better than not applying any dropout for most layers.
However, layers \fc6 and \fc7 are the most interesting ones:
here dropping out $50\%$ random activations decreases the performance substantially, while dropping out $50\%$ least activations only results in a small decrease.
Possibly the exact values of the features in \fc6 and \fc7 do not affect the reconstruction much, but they estimate the importance of different features.

\begin{figure} 
  \footnotesize
  \begin{center}
  \setlength{\tabcolsep}{0.03cm}
  \renewcommand{\arraystretch}{0.5}
  \begin{tabular}{ccccccc}
  \conv5 &
  \raisebox{-.5\height}{\includegraphics[width=0.14\linewidth]{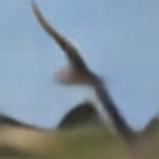}} &
  \raisebox{-.5\height}{\includegraphics[width=0.14\linewidth]{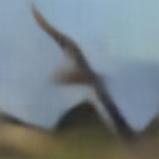}} &
  \raisebox{-.5\height}{\includegraphics[width=0.14\linewidth]{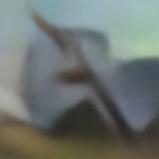}} &
  \raisebox{-.5\height}{\includegraphics[width=0.14\linewidth]{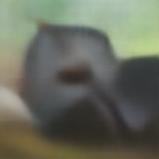}} &
  \raisebox{-.5\height}{\includegraphics[width=0.14\linewidth]{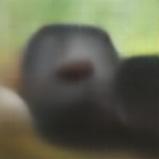}} &
  \raisebox{-.5\height}{\includegraphics[width=0.14\linewidth]{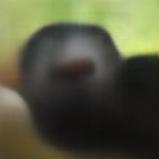}} \vspace*{0.06cm}
  \\
  \fc6 &
  \raisebox{-.5\height}{\includegraphics[width=0.14\linewidth]{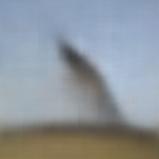}} &
  \raisebox{-.5\height}{\includegraphics[width=0.14\linewidth]{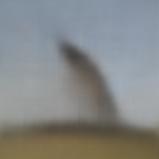}} &
  \raisebox{-.5\height}{\includegraphics[width=0.14\linewidth]{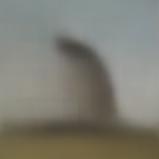}} &
  \raisebox{-.5\height}{\includegraphics[width=0.14\linewidth]{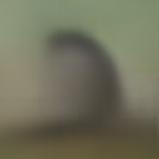}} &
  \raisebox{-.5\height}{\includegraphics[width=0.14\linewidth]{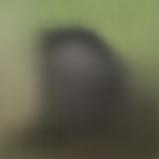}} &
  \raisebox{-.5\height}{\includegraphics[width=0.14\linewidth]{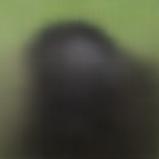}} \vspace*{0.06cm}
  \\
  \fc7 &
  \raisebox{-.5\height}{\includegraphics[width=0.14\linewidth]{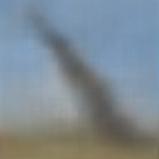}} &
  \raisebox{-.5\height}{\includegraphics[width=0.14\linewidth]{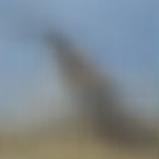}} &
  \raisebox{-.5\height}{\includegraphics[width=0.14\linewidth]{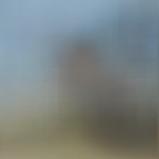}} &
  \raisebox{-.5\height}{\includegraphics[width=0.14\linewidth]{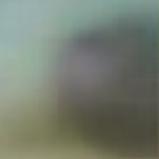}} &
  \raisebox{-.5\height}{\includegraphics[width=0.14\linewidth]{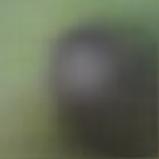}} &
  \raisebox{-.5\height}{\includegraphics[width=0.14\linewidth]{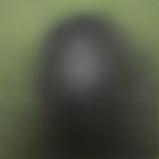}} \vspace*{0.06cm}
  \\
  \fc8 &
  \raisebox{-.5\height}{\includegraphics[width=0.14\linewidth]{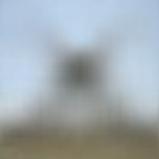}} &
  \raisebox{-.5\height}{\includegraphics[width=0.14\linewidth]{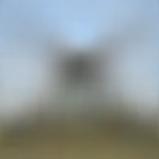}} &
  \raisebox{-.5\height}{\includegraphics[width=0.14\linewidth]{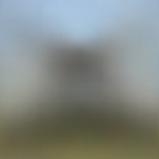}} &
  \raisebox{-.5\height}{\includegraphics[width=0.14\linewidth]{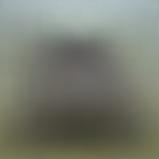}} &
  \raisebox{-.5\height}{\includegraphics[width=0.14\linewidth]{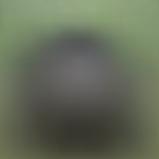}} &
  \raisebox{-.5\height}{\includegraphics[width=0.14\linewidth]{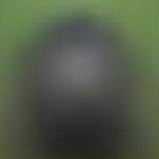}} \vspace*{0.06cm}
  \end{tabular}
  \begin{minipage}{.9\linewidth}
  \centering
  \vspace*{0.4cm}
   \caption{Interpolation between the features of two images.}
\label{fig:interp}
\end{minipage}%
\end{center}
\vspace*{-0.7cm}
\end{figure}

\subsection{Interpolation and random feature vectors}
Another way to analyze the feature representation is by traversing the feature manifold and by observing the corresponding images generated by the reconstruction networks.
We have seen the reconstructions from feature vectors of actual images, but what if a feature vector was not generated from a natural image?
In Figure~\ref{fig:interp} we show reconstructions obtained with our networks when interpolating between feature vectors of two images.
It is interesting to see that interpolating \conv5 features leads to a simple overlay of images, but the behavior of interpolations when reconstructing from \fc6 is very different: images smoothly morph into each other.
More examples, together with the results for autoencoders, are shown in the supplementary material.

Another analysis method is by sampling feature vectors randomly.
Our networks were trained to reconstruct images given their feature representations, but the distribution of the feature vectors is unknown.
Hence, there is no simple principled way to sample from our model.
However, by assuming independence of the features (a very strong and wrong assumption!), we can approximate the distribution of each dimension of the feature vector separately.
To this end we simply computed a histogram of each feature over a set of $4096$ images and sampled from those.
We ensured that the sparsity of the random samples is the same as that of the actual feature vectors.
This procedure led to low contrast images, perhaps because by independently sampling each dimension we did not introduce interactions between the features.
Multiplying the feature vectors by a constant factor $\alpha=2$ increases the contrast without affecting other properties of the generated images.

\begin{figure} 
\small
\begin{center}
\setlength{\tabcolsep}{0.03cm}
\renewcommand{\arraystretch}{0.5}
  \begin{tabular}{cccccccccccccc}
  \conv5 &
  \raisebox{-.5\height}{\includegraphics[width=0.115\linewidth]{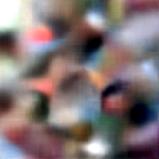}} &
  \raisebox{-.5\height}{\includegraphics[width=0.115\linewidth]{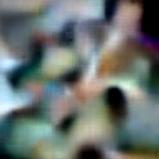}} &
  \raisebox{-.5\height}{\includegraphics[width=0.115\linewidth]{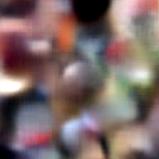}} &
  \raisebox{-.5\height}{\includegraphics[width=0.115\linewidth]{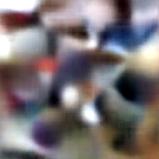}} &
  \raisebox{-.5\height}{\includegraphics[width=0.115\linewidth]{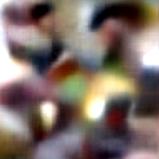}} &
  \raisebox{-.5\height}{\includegraphics[width=0.115\linewidth]{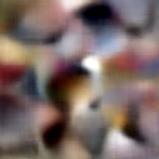}} &
  \raisebox{-.5\height}{\includegraphics[width=0.115\linewidth]{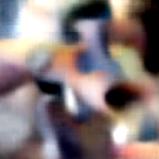}} \vspace*{0.06cm}
  \\
  \fc6 &
  \raisebox{-.5\height}{\includegraphics[width=0.115\linewidth]{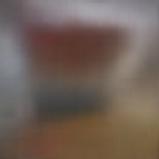}} &
  \raisebox{-.5\height}{\includegraphics[width=0.115\linewidth]{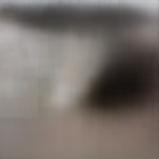}} &
  \raisebox{-.5\height}{\includegraphics[width=0.115\linewidth]{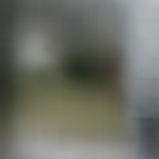}} &
  \raisebox{-.5\height}{\includegraphics[width=0.115\linewidth]{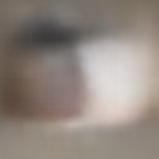}} &
  \raisebox{-.5\height}{\includegraphics[width=0.115\linewidth]{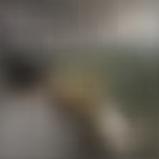}} &
  \raisebox{-.5\height}{\includegraphics[width=0.115\linewidth]{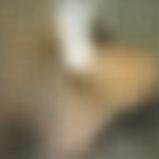}} &
  \raisebox{-.5\height}{\includegraphics[width=0.115\linewidth]{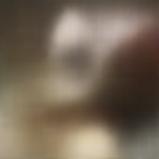}} \vspace*{0.06cm}
  \\
  \fc7 &
  \raisebox{-.5\height}{\includegraphics[width=0.115\linewidth]{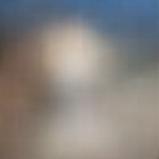}} &
  \raisebox{-.5\height}{\includegraphics[width=0.115\linewidth]{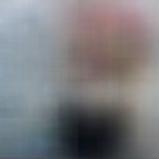}} &
  \raisebox{-.5\height}{\includegraphics[width=0.115\linewidth]{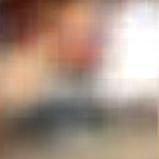}} &
  \raisebox{-.5\height}{\includegraphics[width=0.115\linewidth]{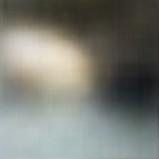}} &
  \raisebox{-.5\height}{\includegraphics[width=0.115\linewidth]{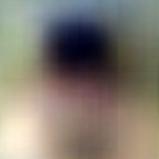}} &
  \raisebox{-.5\height}{\includegraphics[width=0.115\linewidth]{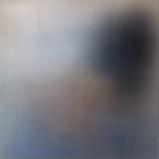}} &
  \raisebox{-.5\height}{\includegraphics[width=0.115\linewidth]{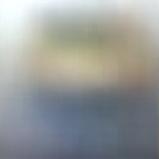}} \vspace*{0.06cm}
  \\
  \fc8 &
  \raisebox{-.5\height}{\includegraphics[width=0.115\linewidth]{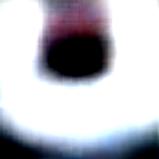}} &
  \raisebox{-.5\height}{\includegraphics[width=0.115\linewidth]{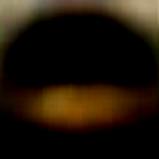}} &
  \raisebox{-.5\height}{\includegraphics[width=0.115\linewidth]{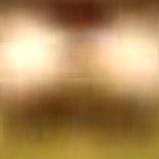}} &
  \raisebox{-.5\height}{\includegraphics[width=0.115\linewidth]{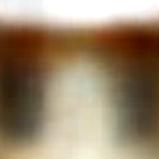}} &
  \raisebox{-.5\height}{\includegraphics[width=0.115\linewidth]{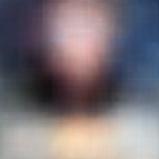}} &
  \raisebox{-.5\height}{\includegraphics[width=0.115\linewidth]{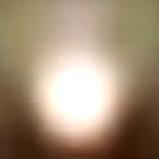}} &
  \raisebox{-.5\height}{\includegraphics[width=0.115\linewidth]{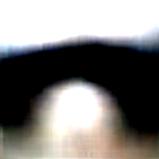}} \vspace*{-0.2cm}
  \end{tabular}
\end{center}
   \caption{Images generated from random feature vectors of top layers of AlexNet.}
\label{fig:gen_examples}
\end{figure}

Random samples obtained this way from four top layers of AlexNet are shown in Figure~\ref{fig:gen_examples}.
No pre-selection was performed.
While samples from \conv5 look much like abstract art, the samples from fully convolutional layers are much more realistic.
This shows that the networks learn a natural image prior that allows them to produce somewhat realistically looking images from random feature vectors.
We found that a much simpler sampling procedure of fitting a single shifted truncated Gaussian to all feature dimensions produces qualitatively very similar images.
These are shown in the supplementary material together with images generated from autoencoders, which look much less like natural images.


%

\section{Conclusions}
We have proposed to invert image representations with up-convolutional networks and have shown that this yields more or less accurate reconstructions of the original images, depending on the level of invariance of the feature representation.
The networks implicitly learn natural image priors which allow the retrieval of information that is obviously lost in the feature representation, such as color or brightness in HOG or SIFT.
The method is very fast at test time and does not require the gradient of the feature representation to be inverted.
Therefore, it can be applied to virtually any image representation.

Application of our method to the representations learned by the AlexNet convolutional network leads do several conclusions:
1) Features from all layers of the network, including the final \fc8 layer, preserve the precise colors and the rough position of objects in the image;
2) In higher layers, almost all information about the input image is contained in the pattern of non-zero activations, not their precise values;
3) In the layer \fc8, most information about the input image is contained in small probabilities of those classes that are not in top-5 network predictions.

\section*{Acknowledgements}
We acknowledge funding by the ERC Starting Grant VideoLearn (279401).  We are grateful to Aravindh Mahendran for sharing with us the reconstructions achieved with the method of Mahendran and Vedaldi~\cite{Mahendran_CVPR2015}. We thank Jost Tobias Springenberg for comments.

{\small
\bibliographystyle{ieee}
\bibliography{dosovits_new}
}

\clearpage

\section*{Supplementary material}

\paragraph{Network architectures}

Table~\ref{tbl:arch_alexnet} shows the architecture of AlexNet.
Tables~\ref{tbl:hog_arch}-\ref{tbl:fc8_arch_supp} show the architectures of networks we used for inverting different features.
After each fully connected and convolutional layer there is always a leaky ReLU nonlinearity.
Networks for inverting HOG and LBP have two streams.
Stream A compresses the input features spatially and accumulates information over large regions.
We found this crucial to get good estimates of the overall brightness of the image.
Stream B does not compress spatially and hence can better preserve fine local details.
At one points the outputs of the two streams are concatenated and processed jointly, denoted by ``J''.
K stands for kernel size, S for stride.

\begin{table*}[] 
\begin{center}
\setlength{\tabcolsep}{0.07cm}
\small{
  \begin{tabular}{|c||c|c|c|c|c|c|c|c|c|c|c|c|c|}
  \hline

layer           & \multicolumn{2}{c|}{\conv1}  & \multicolumn{2}{c|}{\conv2}  & \conv3 & \conv4 & \multicolumn{2}{c|}{\conv5}  &  \multicolumn{2}{c|}{\fc6}   &  \multicolumn{2}{c|}{\fc7}  & \fc8  \\ \hline
processing      & conv1  & mpool1             & conv2 & mpool2              & conv3 & conv4 & conv5 & mpool5              & fc6   & drop6               & fc7   & drop7              & fc8  \\
steps           & relu1  & norm1              & relu2 & norm2               & relu3 & relu4 & relu5 &                     & relu6 &                     & relu7 &                    &      \\ \hline
out size        &  55    &  27                &  27   &  13                 &  13   &   13  &   13  &   6                 &   1   &  1                  &   1   &   1                &  1    \\ \hline
out channels    &  96    &  96                & 256   & 256                 &  384  &  384  & 256   &  256                & 4096  & 4096                & 4096  & 4096               & 1000 \\ \hline
  \end{tabular}
  }\vspace{-5mm}
\end{center}
\caption{Summary of the AlexNet network. Input image size is $227 \times 227$.}
\label{tbl:arch_alexnet}
\end{table*}

\begin{table}[b]
   \begin{center}
   \setlength{\tabcolsep}{0.15cm}
  \resizebox{0.98\linewidth}{!}{%
  \begin{tabular}{|l|cc|cc|c|}
      \hline
      Layer      & Input               & InSize               & K    & S    & OutSize              \\
      \hline                           
      convA1     & HOG                 & $32 \st 32 \st 31$   & $5$  & $2$ & $16 \st 16 \st 256$   \\
      convA2     & convA1              & $16 \st 16 \st 256$  & $5$  & $2$ & $8 \st 8 \st 512$     \\
      convA3     & convA2              & $8 \st 8 \st 512$    & $3$  & $2$ & $4 \st 4 \st 1024$    \\                          
      upconvA1   & convA3              & $4 \st 4 \st 1024$   & $4$  & $2$ & $8 \st 8 \st 512$     \\
      upconvA2   & upconvA1            & $8 \st 8 \st 512$    & $4$  & $2$ & $16 \st 16 \st 256$   \\
      upconvA3   & upconvA2            & $16 \st 16 \st 256$  & $4$  & $2$ & $32 \st 32 \st 128$   \\
      \hline                           
      convB1     & HOG                 & $32 \st 32 \st 31$   & $5$  & $1$ & $32 \st 32 \st 128$   \\
      convB2     & convB1              & $32 \st 32 \st 128$  & $3$  & $1$ & $32 \st 32 \st 128$   \\
      \hline
      convJ1     & \small{\{upconvA3, convB2\}}   & $32 \st 32 \st 256$  & $3$  & $1$ & $32 \st 32 \st 256$   \\
      convJ2     & convJ1              & $32 \st 32 \st 256$  & $3$  & $1$ & $32 \st 32 \st 128$   \\
      upconvJ4   & convJ2              & $32 \st 32 \st 128$  & $4$  & $2$ & $64 \st 64 \st 64$    \\
      upconvJ5   & upconvJ4            & $64 \st 64 \st 64$   & $4$  & $2$ & $128 \st 128 \st 32$  \\
      upconvJ6   & upconvJ5            & $128 \st 128 \st 32$ & $4$  & $2$ & $256 \st 256 \st 3$   \\
      \hline
    \end{tabular}}
  \end{center}
  \vspace*{-0.3cm}
  \caption{Network for reconstructing from HOG features.}
  \label{tbl:hog_arch}
\end{table}

\begin{table}
   \begin{center}
   \setlength{\tabcolsep}{0.15cm}
  \resizebox{0.98\linewidth}{!}{%
  \begin{tabular}{|l|cc|cc|c|}
      \hline
      Layer      & Input               & InSize               & K    & S    & OutSize              \\
      \hline                           
      conv1      & SIFT                & $64 \st 64 \st 133$  & $5$  & $2$ & $32 \st 32 \st 256$   \\
      conv2      & conv1               & $32 \st 32 \st 256$  & $3$  & $2$ & $16 \st 16 \st 512$   \\
      conv3      & conv2               & $16 \st 16 \st 512$  & $3$  & $2$ & $8 \st 8 \st 1024$    \\
      conv4      & conv3               & $8 \st 8 \st 1024$   & $3$  & $2$ & $4 \st 4 \st 2048$    \\
      conv5      & conv4               & $4 \st 4 \st 2048$   & $3$  & $1$ & $4 \st 4 \st 2048$    \\
      conv6      & conv5               & $4 \st 4 \st 2048$   & $3$  & $1$ & $4 \st 4 \st 1024$    \\
      \hline                           
      upconv1    & conv6               & $4 \st 4 \st 1024$   & $4$  & $2$ & $8 \st 8 \st 512$     \\
      upconv2    & upconv1             & $8 \st 8 \st 512$    & $4$  & $2$ & $16 \st 16 \st 256$   \\
      upconv3    & upconv2             & $16 \st 16 \st 256$  & $4$  & $2$ & $32 \st 32 \st 128$   \\
      upconv4    & upconv3             & $32 \st 32 \st 128$  & $4$  & $2$ & $64 \st 64 \st 64$    \\
      upconv5    & upconv4             & $64 \st 64 \st 64$   & $4$  & $2$ & $128 \st 128 \st 32$  \\
      upconv6    & upconv5             & $128 \st 128 \st 32$ & $4$  & $2$ & $256 \st 256 \st 3$   \\
      \hline
    \end{tabular}}
  \end{center}
  \vspace*{-0.3cm}
  \caption{Network for reconstructing from SIFT features.}
  \vspace*{-0.3cm}
  \label{tbl:sift_arch}
\end{table}

\begin{table}
   \begin{center}
   \setlength{\tabcolsep}{0.15cm}
  \resizebox{0.98\linewidth}{!}{%
  \begin{tabular}{|l|cc|cc|c|}
      \hline
      Layer      & Input               & InSize               & K    & S    & OutSize              \\
      \hline                           
      convA1     & LBP                 & $16 \st 16 \st 58$   & $5$  & $2$ & $8 \st 8 \st 256$   \\
      convA2     & convA1              & $8 \st 8 \st 256$    & $5$  & $2$ & $4 \st 4 \st 512$     \\
      convA3     & convA2              & $4 \st 4 \st 512$    & $3$  & $1$ & $4 \st 4 \st 1024$    \\                        
      upconvA1   & convA3              & $4 \st 4 \st 1024$   & $4$  & $2$ & $8 \st 8 \st 512$     \\
      upconvA2   & upconvA1            & $8 \st 8 \st 512$    & $4$  & $2$ & $16 \st 16 \st 256$   \\
      \hline                           
      convB1     & LBP                 & $16 \st 16 \st 58$   & $5$  & $1$ & $16 \st 16 \st 128$   \\
      convB2     & convB1              & $16 \st 16 \st 128$  & $3$  & $1$ & $16 \st 16 \st 128$   \\      
      \hline
      convJ1     & \small{\{upconvA2, convB2\}}   & $16 \st 16 \st 384$  & $3$  & $1$ & $16 \st 16 \st 256$   \\
      convJ2     & convJ1              & $16 \st 16 \st 256$  & $3$  & $1$ & $16 \st 16 \st 128$   \\
      upconvJ3   & convJ2              & $16 \st 16 \st 128$  & $4$  & $2$ & $32 \st 32 \st 128$    \\
      upconvJ4   & upconvJ3            & $32 \st 32 \st 128$  & $4$  & $2$ & $64 \st 64 \st 64$    \\
      upconvJ5   & upconvJ4            & $64 \st 64 \st 64$   & $4$  & $2$ & $128 \st 128 \st 32$  \\
      upconvJ6   & upconvJ5            & $128 \st 128 \st 32$ & $4$  & $2$ & $256 \st 256 \st 3$   \\
      \hline
    \end{tabular}}
  \end{center}
  \vspace*{-0.3cm}
  \caption{Network for reconstructing from LBP features.}
  \vspace*{-0.3cm}
  \label{tbl:lbp_arch}
\end{table}

\begin{table}
   \begin{center}
   \setlength{\tabcolsep}{0.15cm}
  \resizebox{0.98\linewidth}{!}{%
  \begin{tabular}{|l|cc|cc|c|}
      \hline
      Layer      & Input               & InSize               & K    & S    & OutSize              \\
      \hline                           
      conv1      & AlexNet-\conv5      & $6 \st 6 \st 256$    & $3$  & $1$ & $6 \st 6 \st 256$     \\
      conv2      & conv1               & $6 \st 6 \st 256$    & $3$  & $1$ & $6 \st 6 \st 256$     \\
      conv3      & conv2               & $6 \st 6 \st 256$    & $3$  & $1$ & $6 \st 6 \st 256$     \\
      \hline                           
      upconv1    & conv3               & $6 \st 6 \st 256$    & $5$  & $2$ & $12 \st 12 \st 256$   \\
      upconv2    & upconv1             & $12 \st 12 \st 256$  & $5$  & $2$ & $24 \st 24 \st 128$   \\
      upconv3    & upconv2             & $24 \st 24 \st 128$  & $5$  & $2$ & $48 \st 48 \st 64$    \\
      upconv4    & upconv3             & $48 \st 48 \st 64$   & $5$  & $2$ & $96 \st 96 \st 32$    \\
      upconv5    & upconv4             & $96 \st 96 \st 32$   & $5$  & $2$ & $192 \st 192 \st 3$   \\
      \hline
    \end{tabular}}
  \end{center}
  \vspace*{-0.3cm}
  \caption{Network for reconstructing from AlexNet \conv5 features.}
  \vspace*{-0.3cm}
  \label{tbl:conv5_arch}
\end{table}

\begin{table}
   \begin{center}
   \setlength{\tabcolsep}{0.15cm}
  \resizebox{0.98\linewidth}{!}{%
  \begin{tabular}{|l|cc|cc|c|}
      \hline
      Layer      & Input               & InSize               & K    & S    & OutSize              \\
      \hline                           
      fc1        & AlexNet-\fc8        & $1000$               & $-$  & $-$ & $4096$                \\
      fc2        & fc1                 & $4096$               & $-$  & $-$ & $4096$                \\
      fc3        & fc2                 & $4096$               & $-$  & $-$ & $4096$                \\
      reshape    & fc3                 & $4096$               & $-$  & $-$ & $4 \st 4 \st 256$     \\
      \hline                           
      upconv1    & reshape             & $4 \st 4 \st 256$    & $5$  & $2$ & $8 \st 8 \st 256$     \\
      upconv2    & upconv1             & $8 \st 8 \st 256$    & $5$  & $2$ & $16 \st 16 \st 128$   \\
      upconv3    & upconv2             & $16 \st 16 \st 128$  & $5$  & $2$ & $32 \st 32 \st 64$    \\
      upconv4    & upconv3             & $32 \st 32 \st 64$   & $5$  & $2$ & $64 \st 64 \st 32$    \\
      upconv5    & upconv4             & $64 \st 64 \st 32$   & $5$  & $2$ & $128 \st 128 \st 3$   \\
      \hline
    \end{tabular}}
  \end{center}
  \vspace*{-0.3cm}
  \caption{Network for reconstructing from AlexNet \fc8 features.}
  \vspace*{-0.3cm}
  \label{tbl:fc8_arch_supp}
\end{table}

\begin{figure}[]
\begin{center}
\setlength{\tabcolsep}{0.03cm}
\renewcommand{\arraystretch}{0.5}
  \begin{tabular}{cccc}
  Image & HOG our & SIFT our & LBP our
  \\
  {\includegraphics[width=0.24\linewidth]{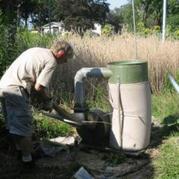}} &
  {\includegraphics[width=0.24\linewidth]{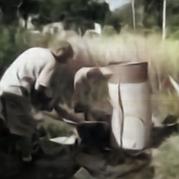}} &
  {\includegraphics[width=0.24\linewidth]{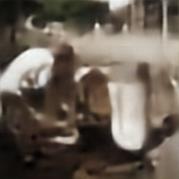}} &
  {\includegraphics[width=0.24\linewidth]{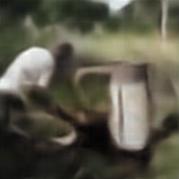}}
  \\
  {\includegraphics[width=0.24\linewidth]{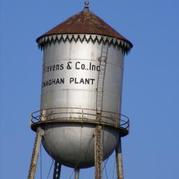}} &
  {\includegraphics[width=0.24\linewidth]{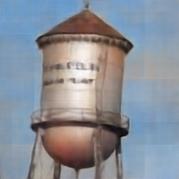}} &
  {\includegraphics[width=0.24\linewidth]{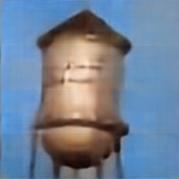}} &
  {\includegraphics[width=0.24\linewidth]{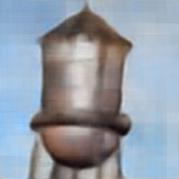}}
  \\
  {\includegraphics[width=0.24\linewidth]{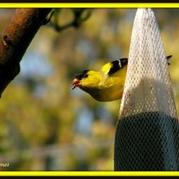}} &
  {\includegraphics[width=0.24\linewidth]{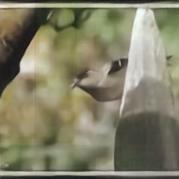}} &
  {\includegraphics[width=0.24\linewidth]{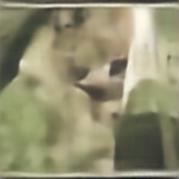}} &
  {\includegraphics[width=0.24\linewidth]{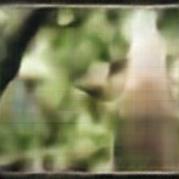}}
  \\
  {\includegraphics[width=0.24\linewidth]{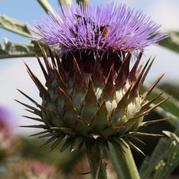}} &
  {\includegraphics[width=0.24\linewidth]{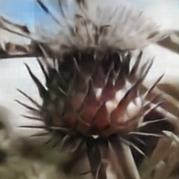}} &
  {\includegraphics[width=0.24\linewidth]{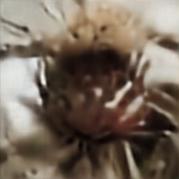}} &
  {\includegraphics[width=0.24\linewidth]{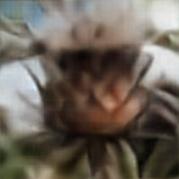}}
    \\
  {\includegraphics[width=0.24\linewidth]{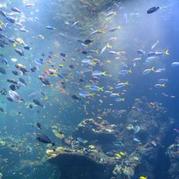}} &
  {\includegraphics[width=0.24\linewidth]{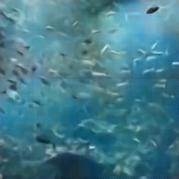}} &
  {\includegraphics[width=0.24\linewidth]{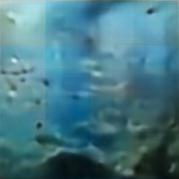}} &
  {\includegraphics[width=0.24\linewidth]{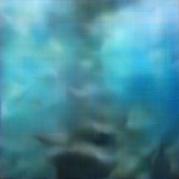}}
  \\
  {\includegraphics[width=0.24\linewidth]{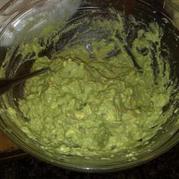}} &
  {\includegraphics[width=0.24\linewidth]{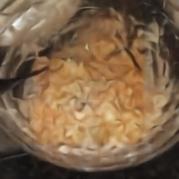}} &
  {\includegraphics[width=0.24\linewidth]{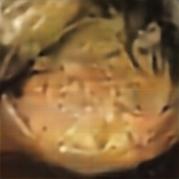}} &
  {\includegraphics[width=0.24\linewidth]{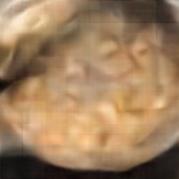}}
  \end{tabular}
\end{center}
\vspace*{-0.5cm}
\caption{Inversion of shallow image representations.}
\label{fig:invert_shallow_supp}
\end{figure}

\paragraph{Shallow features details}
As mentioned, in the paper, for all three methods we use implementations from the \emph{VLFeat} library~\cite{VLfeat} with the default settings.
We use the Felzenszwalb et al. version of HOG with cell size $8$.
For SIFT we used $3$ levels per octave, the first octave was $0$ (corresponding to full resolution), the number of octaves was set automatically, effectively searching keypoints of all possible sizes.

The LBP version we used works with $3\times 3$ pixel neighborhoods.
Each of the $8$ non-central bits is equal to one if the corresponding pixel is brighter than the central one. 
All possible $256$ patterns are quantized into $58$ patterns.
These include $56$ patterns with exactly one transition from $0$ to $1$ when going around the central pixel, plus one quantized pattern comprising two uniform patterns, plus one quantized pattern containing all other patterns.
The quantized LBP patterns are then grouped into local histograms over cells of $16 \times 16$ pixels.

\paragraph{Experiments: shallow representations}
Figure~\ref{fig:invert_shallow_supp} shows several images and their reconstructions from HOG, SIFT and LBP. 
HOG allows for the best reconstruction, SIFT slightly worse, LBP yet slightly worse.
Colors are often reconstructed correctly, but sometimes are wrong, for example in the last row.
Interestingly, all network typically agree on estimated colors.

\begin{figure*}[!h] 
\begin{center}
\setlength{\tabcolsep}{0.03cm}
\renewcommand{\arraystretch}{0.5}
\begin{tabular}{ccccccccc}
  Image & \conv1 & \conv2 & \conv3 & \conv4 & \conv5 & \fc6 & \fc7 & \fc8
  \\
  {\includegraphics[width=0.105\linewidth]{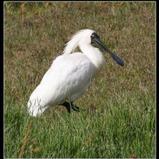}} &
  {\includegraphics[width=0.105\linewidth]{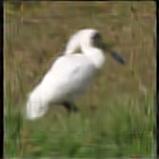}} &
  {\includegraphics[width=0.105\linewidth]{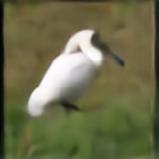}} &
  {\includegraphics[width=0.105\linewidth]{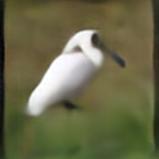}} &
  {\includegraphics[width=0.105\linewidth]{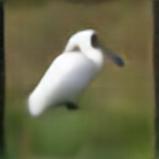}} &
  {\includegraphics[width=0.105\linewidth]{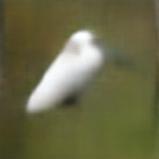}} &
  {\includegraphics[width=0.105\linewidth]{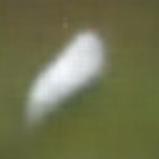}} &
  {\includegraphics[width=0.105\linewidth]{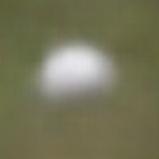}} &
  {\includegraphics[width=0.105\linewidth]{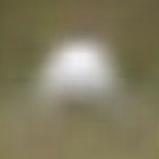}}
  \\
  {\includegraphics[width=0.105\linewidth]{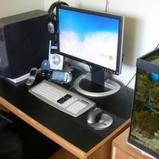}} &
  {\includegraphics[width=0.105\linewidth]{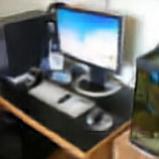}} &
  {\includegraphics[width=0.105\linewidth]{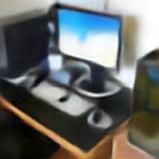}} &
  {\includegraphics[width=0.105\linewidth]{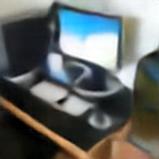}} &
  {\includegraphics[width=0.105\linewidth]{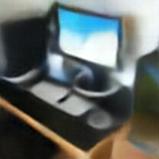}} &
  {\includegraphics[width=0.105\linewidth]{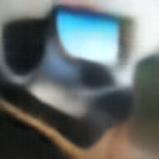}} &
  {\includegraphics[width=0.105\linewidth]{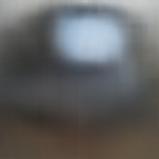}} &
  {\includegraphics[width=0.105\linewidth]{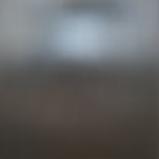}} &
  {\includegraphics[width=0.105\linewidth]{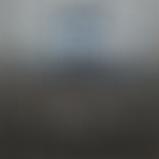}}
  \\
\end{tabular}
\end{center}
   \caption{Reconstructions from different layers of AlexNet.}
\label{fig:recon_examples_supp}
\end{figure*}

\begin{figure*} 
\footnotesize
\begin{minipage}{.35\textwidth} 
\begin{center}
\setlength{\tabcolsep}{0.03cm}
\renewcommand{\arraystretch}{0.5}
  \begin{tabular}{cccc}
  Image & all & top5 & notop5 
  \\
  {\includegraphics[width=0.21\linewidth]{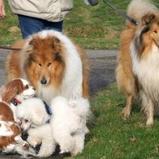}} &
  {\includegraphics[width=0.21\linewidth]{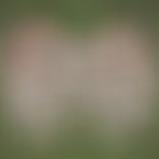}} &
  {\includegraphics[width=0.21\linewidth]{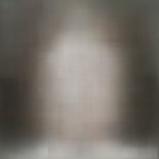}} &
  {\includegraphics[width=0.21\linewidth]{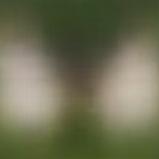}} 
  \\ 
  {\includegraphics[width=0.21\linewidth]{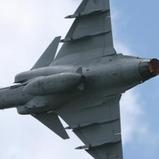}} &
  {\includegraphics[width=0.21\linewidth]{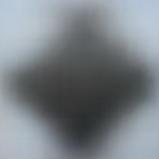}} &
  {\includegraphics[width=0.21\linewidth]{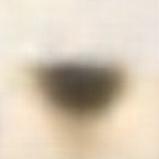}} &
  {\includegraphics[width=0.21\linewidth]{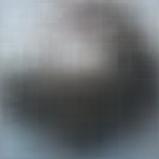}} 
  \\ 
  {\includegraphics[width=0.21\linewidth]{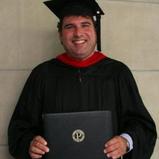}} &
  {\includegraphics[width=0.21\linewidth]{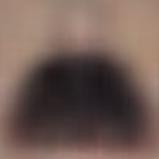}} &
  {\includegraphics[width=0.21\linewidth]{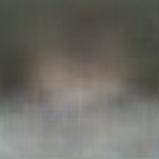}} &
  {\includegraphics[width=0.21\linewidth]{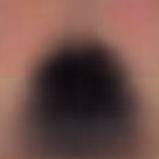}} 
  \end{tabular}
\end{center}
\begin{minipage}{.95\textwidth}
   \caption{Left to right: input image, reconstruction from fc8, reconstruction from $5$ largest activations in \fc8, reconstruction from all \fc8 activations except $5$ largest ones.}
   \label{fig:top5_notop5}
   \end{minipage}
\end{minipage}
\begin{minipage}{.65\textwidth}
\begin{center}
 \hspace{-0.4cm}
\setlength{\tabcolsep}{0.03cm}
\renewcommand{\arraystretch}{0.5}
  \begin{tabular}{ccccccccc}
  \;\;\fc6 &
  \raisebox{-.5\height}{\includegraphics[width=0.11\linewidth]{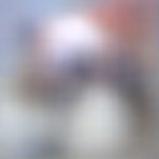}} &
  \raisebox{-.5\height}{\includegraphics[width=0.11\linewidth]{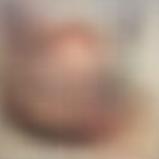}} &
  \raisebox{-.5\height}{\includegraphics[width=0.11\linewidth]{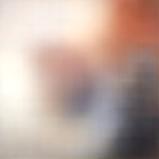}} &
  \raisebox{-.5\height}{\includegraphics[width=0.11\linewidth]{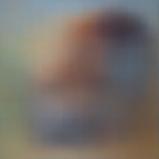}} &
  \raisebox{-.5\height}{\includegraphics[width=0.11\linewidth]{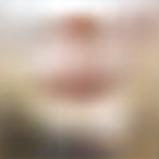}} &
  \raisebox{-.5\height}{\includegraphics[width=0.11\linewidth]{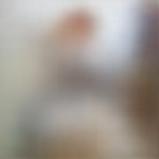}} &
  \raisebox{-.5\height}{\includegraphics[width=0.11\linewidth]{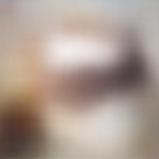}} &
  \raisebox{-.5\height}{\includegraphics[width=0.11\linewidth]{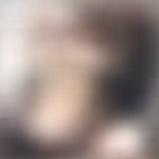}} \vspace*{0.06cm}
  \\
  \;\;\fc7 &
  \raisebox{-.5\height}{\includegraphics[width=0.11\linewidth]{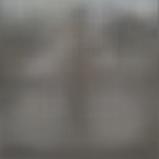}} &
  \raisebox{-.5\height}{\includegraphics[width=0.11\linewidth]{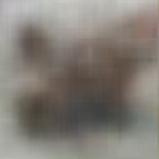}} &
  \raisebox{-.5\height}{\includegraphics[width=0.11\linewidth]{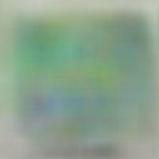}} &
  \raisebox{-.5\height}{\includegraphics[width=0.11\linewidth]{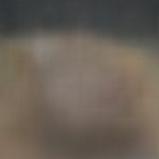}} &
  \raisebox{-.5\height}{\includegraphics[width=0.11\linewidth]{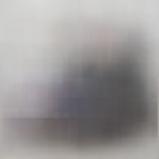}} &
  \raisebox{-.5\height}{\includegraphics[width=0.11\linewidth]{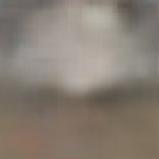}} &
  \raisebox{-.5\height}{\includegraphics[width=0.11\linewidth]{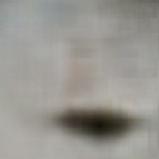}} &
  \raisebox{-.5\height}{\includegraphics[width=0.11\linewidth]{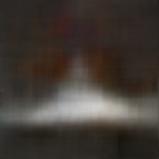}} \vspace*{0.06cm}
  \\
\end{tabular}
\end{center}\hspace{0.4cm}
\begin{minipage}{.93\textwidth}
   \caption{Reconstructions from single neuron activations in the fully connected layers of AlexNet. The \fc8 neurons correspond to classes, left to right: kite, convertible, desktop computer, school bus, street sign, soup bowl, bell pepper, soccer ball.}
   \label{fig:gen_single_neurons}
   \end{minipage}
\end{minipage}
\end{figure*}

\paragraph{Experiments: AlexNet}
We show here several additional figures similar to ones from the main paper. 
Reconstructions from different layers of AlexNet are shown in Figure~\ref{fig:recon_examples_supp}\,.
Figure~\ref{fig:top5_notop5} shows results illustrating the 'dark knowledge' hypothesis, similar to Figure 8 from the main paper.
We reconstruct from all \fc8 features, as well as from only 5 largest ones or all except the 5 largest ones. 
It turns out that the top 5 activations are not very important.

Figure~\ref{fig:gen_single_neurons} shows images generated by activating single neurons in different layers and setting all other neurons to zero.
Particularly interpretable are images generated this way from \fc8. 
Every \fc8 neuron corresponds to a class. Hence the image generated from the activation of, say, ``apple'' neuron, could be expected to be a stereotypical apple.
What we observe looks rather like it might be the average of all images of the class.
For some classes the reconstructions are somewhat interpretable, for others -- not so much.

\begin{figure*} 
\begin{center}
\setlength{\tabcolsep}{0.03cm}
\renewcommand{\arraystretch}{0.5}
  \begin{tabular}{cccccccccc}
     & Image & \conv1 & \conv2 & \conv3 & \conv4 & \conv5 & \fc6 & \fc7 & \fc8
  \\
  Our &
  \raisebox{-.5\height}{\includegraphics[width=0.1\linewidth]{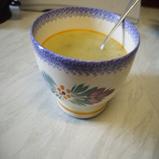}} &
  \raisebox{-.5\height}{\includegraphics[width=0.1\linewidth]{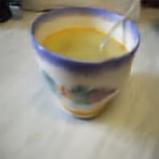}} &
  \raisebox{-.5\height}{\includegraphics[width=0.1\linewidth]{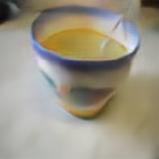}} &
  \raisebox{-.5\height}{\includegraphics[width=0.1\linewidth]{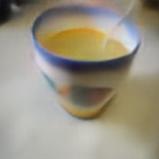}} &
  \raisebox{-.5\height}{\includegraphics[width=0.1\linewidth]{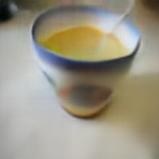}} &
  \raisebox{-.5\height}{\includegraphics[width=0.1\linewidth]{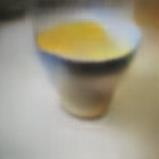}} &
  \raisebox{-.5\height}{\includegraphics[width=0.1\linewidth]{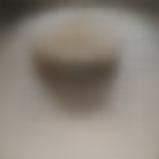}} &
  \raisebox{-.5\height}{\includegraphics[width=0.1\linewidth]{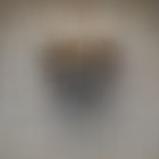}} &
  \raisebox{-.5\height}{\includegraphics[width=0.1\linewidth]{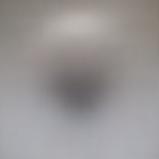}} \vspace*{0.1cm} 
  \\
\end{tabular}
\end{center}
   \caption{Reconstructions from different layers of AlexNet with our method and ~\cite{Mahendran_CVPR2015}.}
\label{fig:recon_compare_supp}
\end{figure*}

Qualitative comparison of reconstructions with our method to the reconstructions of~\cite{Mahendran_CVPR2015} and the results with AlexNet-based autoencoders is given in Figure~\ref{fig:recon_compare_supp}\,.

\begin{figure*} 
\begin{center}
  \setlength{\tabcolsep}{0.03cm}
  \renewcommand{\arraystretch}{0.5}
  \begin{tabular}{cccccccccccccc}
  \conv4 &
  \raisebox{-.5\height}{\includegraphics[width=0.07\linewidth]{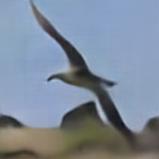}} &
  \raisebox{-.5\height}{\includegraphics[width=0.07\linewidth]{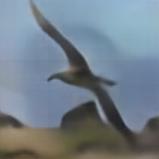}} &
  \raisebox{-.5\height}{\includegraphics[width=0.07\linewidth]{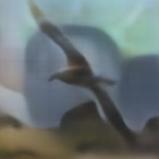}} &
  \raisebox{-.5\height}{\includegraphics[width=0.07\linewidth]{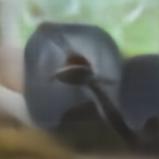}} &
  \raisebox{-.5\height}{\includegraphics[width=0.07\linewidth]{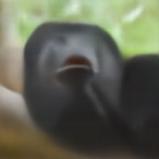}} &
  \raisebox{-.5\height}{\includegraphics[width=0.07\linewidth]{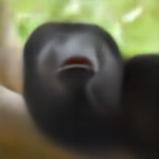}} &
  \qquad\qquad &
  \raisebox{-.5\height}{\includegraphics[width=0.07\linewidth]{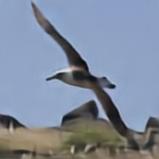}} &
  \raisebox{-.5\height}{\includegraphics[width=0.07\linewidth]{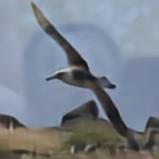}} &
  \raisebox{-.5\height}{\includegraphics[width=0.07\linewidth]{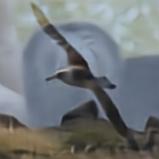}} &
  \raisebox{-.5\height}{\includegraphics[width=0.07\linewidth]{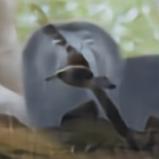}} &
  \raisebox{-.5\height}{\includegraphics[width=0.07\linewidth]{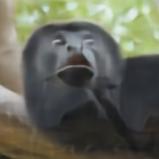}} &
  \raisebox{-.5\height}{\includegraphics[width=0.07\linewidth]{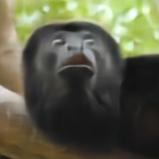}} \vspace*{0.06cm}
  \\
\end{tabular}
   \caption{Interpolation between the features of two images. Left: AlexNet weights fixed, right: autoencoder.}
\label{fig:interp_ae}
\end{center}
\end{figure*}

\begin{figure*} 
\begin{center}
  \setlength{\tabcolsep}{0.03cm}
  \renewcommand{\arraystretch}{0.5}
  \begin{tabular}{cccccccccccccc}
  \conv4 &
  \raisebox{-.5\height}{\includegraphics[width=0.07\linewidth]{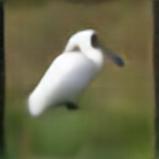}} &
  \raisebox{-.5\height}{\includegraphics[width=0.07\linewidth]{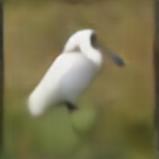}} &
  \raisebox{-.5\height}{\includegraphics[width=0.07\linewidth]{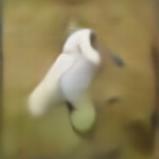}} &
  \raisebox{-.5\height}{\includegraphics[width=0.07\linewidth]{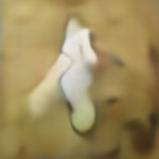}} &
  \raisebox{-.5\height}{\includegraphics[width=0.07\linewidth]{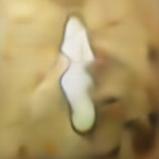}} &
  \raisebox{-.5\height}{\includegraphics[width=0.07\linewidth]{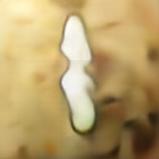}} &
  \qquad\; &
  \raisebox{-.5\height}{\includegraphics[width=0.07\linewidth]{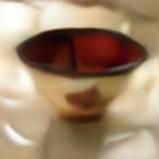}} &
  \raisebox{-.5\height}{\includegraphics[width=0.07\linewidth]{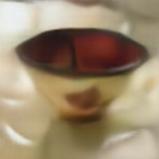}} &
  \raisebox{-.5\height}{\includegraphics[width=0.07\linewidth]{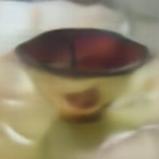}} &
  \raisebox{-.5\height}{\includegraphics[width=0.07\linewidth]{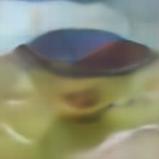}} &
  \raisebox{-.5\height}{\includegraphics[width=0.07\linewidth]{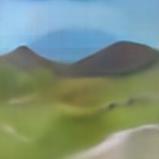}} &
  \raisebox{-.5\height}{\includegraphics[width=0.07\linewidth]{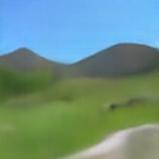}} \vspace*{0.06cm}
  \\
\end{tabular}
   \caption{More interpolations between the features of two images with fixed AlexNet weights.}
\label{fig:interp_supp}
\end{center}
\end{figure*}

Reconstructions from feature vectors obtained by interpolating between feature vectors of two images are shown in Figure~\ref{fig:interp_ae}\,, both for fixed AlexNet and autoencoder training.
More examples of such interpolations with fixed AlexNet are shown in Figure~\ref{fig:interp_supp}\,.

\begin{figure*} 
\small
\begin{center}
\setlength{\tabcolsep}{0.03cm}
\renewcommand{\arraystretch}{0.5}
  \begin{tabular}{cccccccccccccc}
  \conv5 &
  \raisebox{-.5\height}{\includegraphics[width=0.065\linewidth]{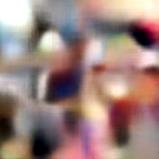}} &
  \raisebox{-.5\height}{\includegraphics[width=0.065\linewidth]{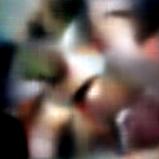}} &
  \raisebox{-.5\height}{\includegraphics[width=0.065\linewidth]{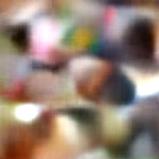}} &
  \raisebox{-.5\height}{\includegraphics[width=0.065\linewidth]{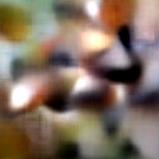}} &
  \raisebox{-.5\height}{\includegraphics[width=0.065\linewidth]{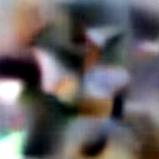}} &
  \raisebox{-.5\height}{\includegraphics[width=0.065\linewidth]{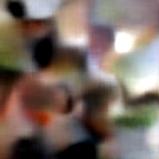}} &
  \raisebox{-.5\height}{\includegraphics[width=0.065\linewidth]{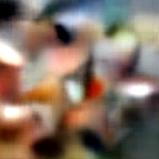}} &
  \raisebox{-.5\height}{\includegraphics[width=0.065\linewidth]{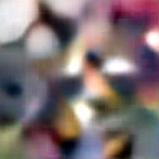}} &
  \raisebox{-.5\height}{\includegraphics[width=0.065\linewidth]{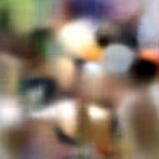}} &
  \raisebox{-.5\height}{\includegraphics[width=0.065\linewidth]{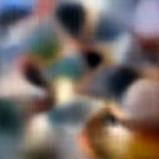}} &
  \raisebox{-.5\height}{\includegraphics[width=0.065\linewidth]{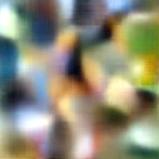}} &
  \raisebox{-.5\height}{\includegraphics[width=0.065\linewidth]{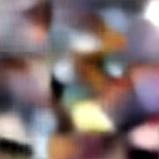}} &
  \raisebox{-.5\height}{\includegraphics[width=0.065\linewidth]{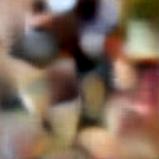}} \vspace*{0.06cm}
  \\
\end{tabular}
\end{center}
   \caption{Images generated from random feature vectors of top layers of AlexNet with the simpler truncated Gaussian distribution (see section 5.5 of the main paper).}
\label{fig:gen_examples_gaussian}
\end{figure*}

\begin{figure*} 
\small
\begin{center}
\setlength{\tabcolsep}{0.03cm}
\renewcommand{\arraystretch}{0.5}
  \begin{tabular}{cccccccccccccc}
  \conv5 &
  \raisebox{-.5\height}{\includegraphics[width=0.065\linewidth]{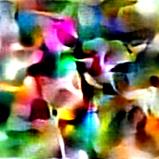}} &
  \raisebox{-.5\height}{\includegraphics[width=0.065\linewidth]{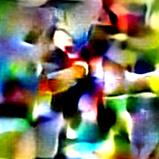}} &
  \raisebox{-.5\height}{\includegraphics[width=0.065\linewidth]{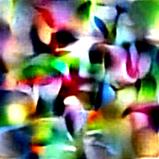}} &
  \raisebox{-.5\height}{\includegraphics[width=0.065\linewidth]{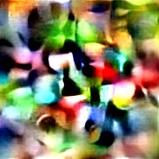}} &
  \raisebox{-.5\height}{\includegraphics[width=0.065\linewidth]{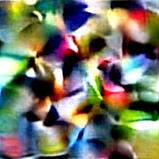}} &
  \raisebox{-.5\height}{\includegraphics[width=0.065\linewidth]{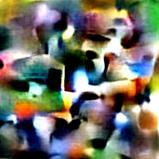}} &
  \raisebox{-.5\height}{\includegraphics[width=0.065\linewidth]{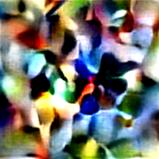}} &
  \raisebox{-.5\height}{\includegraphics[width=0.065\linewidth]{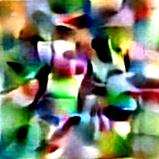}} &
  \raisebox{-.5\height}{\includegraphics[width=0.065\linewidth]{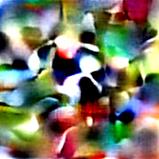}} &
  \raisebox{-.5\height}{\includegraphics[width=0.065\linewidth]{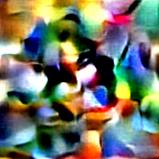}} &
  \raisebox{-.5\height}{\includegraphics[width=0.065\linewidth]{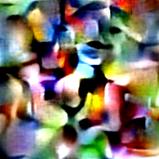}} &
  \raisebox{-.5\height}{\includegraphics[width=0.065\linewidth]{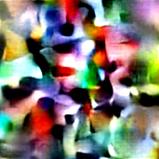}} &
  \raisebox{-.5\height}{\includegraphics[width=0.065\linewidth]{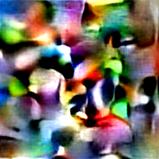}} \vspace*{0.06cm}
  \\
\end{tabular}
\end{center}
   \caption{Images generated from random feature vectors of top layers of AlexNet-based autoencoders with the histogram-based distribution (see section 5.5 of the main paper).}
\label{fig:gen_examples_ae}
\end{figure*}

\begin{figure*} 
\small
\begin{center}
\setlength{\tabcolsep}{0.03cm}
\renewcommand{\arraystretch}{0.5}
  \begin{tabular}{cccccccccccccc}
  \conv5 &
  \raisebox{-.5\height}{\includegraphics[width=0.065\linewidth]{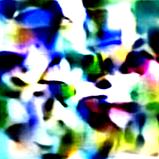}} &
  \raisebox{-.5\height}{\includegraphics[width=0.065\linewidth]{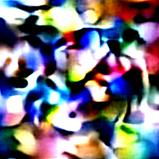}} &
  \raisebox{-.5\height}{\includegraphics[width=0.065\linewidth]{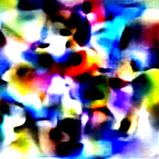}} &
  \raisebox{-.5\height}{\includegraphics[width=0.065\linewidth]{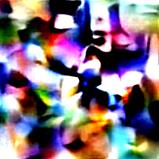}} &
  \raisebox{-.5\height}{\includegraphics[width=0.065\linewidth]{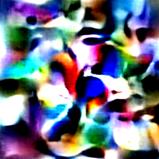}} &
  \raisebox{-.5\height}{\includegraphics[width=0.065\linewidth]{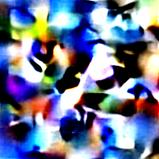}} &
  \raisebox{-.5\height}{\includegraphics[width=0.065\linewidth]{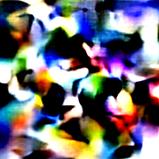}} &
  \raisebox{-.5\height}{\includegraphics[width=0.065\linewidth]{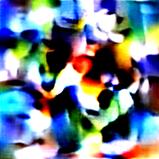}} &
  \raisebox{-.5\height}{\includegraphics[width=0.065\linewidth]{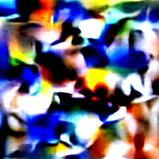}} &
  \raisebox{-.5\height}{\includegraphics[width=0.065\linewidth]{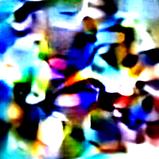}} &
  \raisebox{-.5\height}{\includegraphics[width=0.065\linewidth]{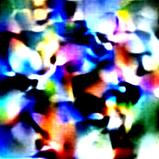}} &
  \raisebox{-.5\height}{\includegraphics[width=0.065\linewidth]{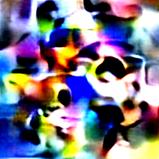}} &
  \raisebox{-.5\height}{\includegraphics[width=0.065\linewidth]{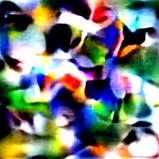}} \vspace*{0.06cm}
  \\
\end{tabular}
\end{center}
   \caption{Images generated from random feature vectors of top layers of AlexNet-based autoencoders with the simpler truncated Gaussian distribution (see section 5.5 of the main paper).}
\label{fig:gen_examples_ae_gaussian}
\end{figure*}

As described in section 5.5 of the main paper, we tried two different distributions for sampling random feature activations: a histogram-based and a truncated Gaussian.
Figure~\ref{fig:gen_examples_gaussian} shows the results with fixed AlexNet network and truncated Gaussian distribution.
Figures~\ref{fig:gen_examples_ae} and~\ref{fig:gen_examples_ae_gaussian} show images generated with autoencoder-trained networks.
Note that images generated from autoencoders look much less realistic than images generated with a network with fixed AlexNet weights.
This indicates that reconstructing from AlexNet features requires a strong natural image prior.

\end{document}